\documentclass[12pt]{article}
\usepackage[utf8]{inputenc}
\DeclareUnicodeCharacter{2212}{-}
\usepackage{amsmath}
\usepackage{amsfonts}
\usepackage{amssymb}
\usepackage{array}
\usepackage{upgreek}
\usepackage{longtable}
\usepackage{booktabs}
\usepackage[toc,page]{appendix}
\usepackage{caption}
\usepackage{subcaption}
\usepackage{graphicx}
\usepackage{tikz}
\usepackage{multicol}
\usepackage{comment}
\usepackage[backend=biber, natbib=true, citestyle=authoryear-comp, sorting=ynt, maxcitenames=2, mincitenames=1, maxbibnames=99,uniquelist=false]{biblatex}
\usepackage[colorlinks=true,linkcolor=blue,citecolor=blue]{hyperref}

\DeclareFieldFormat{citehyperref}{%
  \DeclareFieldAlias{bibhyperref}{noformat}
  \bibhyperref{#1}}

\savebibmacro{cite}

\renewbibmacro*{cite}{%
  \printtext[citehyperref]{%
    \restorebibmacro{cite}%
    \usebibmacro{cite}}}

\begin{document}


\begin{center}
{\Large
	{\sc  MMM: Clustering Multivariate Longitudinal Mixed-type Data}
}
\bigskip

Francesco Amato $^{1}$, Julien Jacques $^{1}$
\bigskip

{\it
$^{1}$ Univ Lyon, Univ Lyon 2, ERIC, Lyon.\\
\{\text{francesco.amato, julien.jacques}\}@univ-lyon2.fr\\ 
}
\end{center}
\bigskip


\medskip

{\bf Abstract.} Multivariate longitudinal data of mixed-type are increasingly collected in many science domains. However, algorithms to cluster this kind of data remain scarce, due to the challenge to simultaneously model the within- and between-time dependence structures for multivariate data of mixed kind. We introduce the Mixture of Mixed-Matrices (MMM) model: reorganizing the data in a three-way structure and assuming that the non-continuous variables are observations of underlying latent continuous variables, the model relies on a mixture of matrix-variate normal distributions to perform clustering in the latent dimension. The MMM model is thus able to handle continuous, ordinal, binary, nominal and count data and to concurrently model the heterogeneity, the association among the responses and the temporal dependence structure in a parsimonious way and without assuming conditional independence. The inference is carried out through an MCMC-EM algorithm, which is detailed. An evaluation of the model through synthetic data shows its inference abilities. A real-world application on financial data is presented. \\

 Mixed-type multivariate longitudinal data. Three-way data. Mixture models. Matrix-variate Gaussians.

\section{Context}
Multivariate longitudinal data of mixed-type are increasingly collected in many science domains. For example, in social sciences studies are often based on questionnaires encompassing different type of answers completed by participants multiple times. In physical sciences, phenomena are often measured repeatedly with different types of measurements.\\
However, the statistical analysis of these data is far from simple, for several reasons. First, the collected data are often of different typology, such as continuous, categorical or count data. The analysis of such mixed-type data is a current research problem in statistics and machine learning \citep{Ahmad2019Mar}.
The second scientific obstacle is the modeling of the temporal trajectory.\\

In this work we aim at providing a tool to perform clustering on multivariate longitudinal mixed-type data. Probabilistic (or model-based) clustering offers the advantage of clearly stating the assumptions behind the clustering algorithm, and allows cluster analysis to benefit from the inferential framework of statistics to address some of the practical questions arising when performing clustering \citep{bouveyron2019model}. 

\subsection{Related work}
\label{sec:related_work}
While several approaches exist for the clustering longitudinal and mixed-type data separately, the body of literature remains comparatively sparse when they are to be dealt with simultaneously. In the following, we will present a brief overlook to the main methods to cluster mixed data, longitudinal data and mixed longitudinal data.\\

Although many data sets contain mixed-type data, few mixture models can manage these data \citep{Hunt2011Jul} due to the shortage of multivariate distributions able to handle them. Clustering with mixed-type data have received a large attention in the last decade from the researcher in statistics and machine learning. The latent class model \citep{everitt} is frequently used, and it assumes that the variables are conditionally independent upon the cluster membership. Consequently, the joint probability distribution function (pdf) of the variables of different types is obtained by the product of the pdfs of each individual variable. However, when the variables are inherently correlated in a cluster, this model is not suitable. To overcome this issue, \cite{marbac2017model} used Gaussian copulas to loosen conditional independence assumptions.  However, the authors note that model complexity increases promptly with the number of variables. Moreover, it is not easily interpretable by practitioners without statistical training. More recently, \cite{Hermes2024Jan} proposed a similar approach by using copulas in the context of graphical models, which were already extended for use for mixed-type data by \cite{Cheng2017Apr}. In \cite{Selosse2020Apr}, another model-based approach for ordinal, nominal, integer and continuous data is proposed, on the basis of conditional independence assumption and with the particularity of creating clusters of variables as well as clusters of individuals (co-clustering).\\
Another way to address the issues of mixed-type data is to see some variables as the manifestation of latent variables. For example, in \cite{clustmd}, the clustMD model considers continuous and categorical data (nominal and ordinal) and assumes that a categorical variable is the representation of an underlying latent continuous variable. Then, it is assumed that the continuous variables (observed and unobserved) follow a multivariate Gaussian mixture model. This model is further developed to address sparsity by \cite{Choi2023Jul}.\\

Modeling longitudinal data poses a different kind of challenge than mixed-type data, as the grouping has to account for the similarity of individual trajectories which disrupt the independence assumption among observations. Additionally, this kind of data introduces the issue of dealing with time, often with sparse observations that makes unsuitable the use of models coming from the domains such as functional data, time series and  Gaussian processes. In order to bypass these problems, some authors preferred to focus on geometric non-parametric clustering algorithms, as done by \cite{Bruckers2016Jul} with an idea based on k-means clustering and by \cite{Zhou2023Jul} with hierarchical clustering, among others. For parametric methods, a well established manner to model longitudinal data is through mixed-effects models. We refer to \cite{Gad2012} for an overview and to the related work section of \cite{Hui2024Oct} for the most recent advancements. The main issues with this kind of models are the over-parametrization and the computational burden that often arises with it.\\

Another approach to clustering longitudinal data  that gained traction in the last decade consists in arranging the data in a three-way format and modeling them through a matrix-variate mixture model. This approach offers the advantage of accounting for the overall time-behavior, grouping together the units that have a similar pattern across and within time. While not being new \citep{basford1985mixture}, matrix-variate distributions have recently gained attention, and mixtures of matrix-normals (MMN) have been developed in both frequentist \citep{Viroli2011Oct} and Bayesian \citep{Viroli2011Dec} frameworks. These models represent a natural extension of the multivariate normal mixtures to account for temporal (or even spatial) dependencies, and have the advantage of being also relatively easy to estimate by means of EM algorithm (a nice short description of the EM application to MNN is provided in §2.1 of \cite{wang2020variable}). 
In addition, in the context of linear mixed models with discrete individual random intercepts to analyze longitudinal continuous data, \cite{Anderlucci2015Jun} proposed Covariance Pattern Mixture Model (CPMM) which, by leveraging three-way data structures, does not require the usual local independence assumption. This model can be seen as an extension of the proposal of \cite{Brendan2010} in the multivariate context.
More recently, The aim was primarily to address the limitations of the MMN model in terms of parsimony and inability to cope with skewed and/or leptokurtic clusters. in \cite{gallaugher2018finite} and \cite{Melnykov2018Sep,melnykov2019studying} extensions for non-normal skewed cases have been proposed and applied. However, matrix-variate models suffer from over-parametrization that leads to estimation issues. To overcome this issue a more parsimonious model \citep{Sarkar2020Feb} and a new R package \citep{Zhu2022Mar} has been proposed.
In addition, \cite{Cappozzo2024Dec} proposed a lasso-type penalization to account for sparsity.
Despite their efficacy, up to now these methods have generally only been applied to continuous data.\\ 
More recently, \cite{Amato2024Apr} proposed a method to cluster longitudinal ordinal data by assuming an underlying mixture of matrix-variate distributions.\\

A significant advancement in matrix-variate longitudinal modeling has been the introduction of hidden Markov models (HMMs) that allow for time-varying cluster membership. \cite{Tomarchio2022} introduced parsimonious HMMs for matrix-variate balanced longitudinal data using eigen decomposition to address overparameterization. This framework was extended by \cite{Tomarchio2024} to include regression components with fixed and random covariates, and further enhanced by \cite{Tomarchio2025} with heavy-tailed distributions to improve robustness against outlying observations.\\

Finally, looking at mixed-type longitudinal data, one main methodology to deal with such data lies in the framework of discrete (time-constant or varying) random intercepts for modeling heterogeneity, that includes mixture random effect models for longitudinal data extended to deal with multivariate and mixed outcomes by \cite{Proust-Lima2013Nov} and growth mixture models (\cite{Ram2009Oct}), where individuals are grouped in classes having a specific growth structure variability. These approaches are similar in that they model the change over time at both the population level and the individual level using random effects (or latent variables).
In \cite{Komarek2013Mar} the authors rely on a multivariate extension of the classical generalized linear mixed model where a mixture distribution is additionally assumed for random effects. \cite{Vavra2023Jun} extend this model presenting a statistical model for joint modeling of mixed-type longitudinal data, while performing unsupervised clustering with respect to different covariate patterns. However, nominal (polytomous) variables are not taken into account in neither of the papers and time-dependent information is neglected. This work is expanded and improved in \cite{Vavra2024Feb}. In \cite{cagnone2018multivariate} the authors extended the latent class model to take into account time evolution by means of latent Markov variable  (\cite{bartolucci2012latent}) to model longitudinal binary and ordinal data on alcohol use disorder.\\
In a model-based clustering perspective, \cite{DelaCruz-Mesia2008Jan} proposed a mixture of hierarchical nonlinear models for describing nonlinear relationships across time. \cite{Manrique-Vallier2014Dec} introduced a
clustering strategy based on a mixed membership framework for analyzing discrete multivariate longitudinal data.\\

\subsection{Preliminaries}
\label{sec:prel}
Let $Z \sim \mathcal{MN}_{(J\times T)}(M,\Phi,\Sigma)$, that is a matrix-variate normal distribution where $M \in \mathbb{R}^{J\times T}$ is the matrix of means, $\Phi \in \mathbb{R}^{T \times T}$ is a covariance matrix containing the variances and covariances between the $T$ occasions or times and $\Sigma \in \mathbb{R}^{J \times J}$ is the covariance matrix containing the variances and covariances of the $J$ variables. The matrix-normal probability density function is given by
\begin{equation}
f(Z|M,\Phi,\Sigma) = (2\pi)^{-\frac{TJ}{2}}|\Phi|^{-\frac{J}{2}}|\Sigma|^{-\frac{T}{2}}
\exp\left\{-\frac{1}{2} \text{tr}[ \Sigma^{-1}(Z-M)\Phi^{-1}(Z-M)^{\top}] \right\}.
\end{equation}
The matrix-normal distribution represents a natural extension of the multivariate normal distribution, since if $Z \sim \mathcal{MN}_{(J\times T)}(M,\Phi,\Sigma)$, then $\text{vec}(Z) \sim \mathcal{N}_{JT}(\text{vec}(M),\Phi \otimes \Sigma)$, that is the multivariate normal distribution of dimension $JT$, where $\text{vec}(.)$ is the vectorization operator, that is the function mapping from a $J \times T$ matrix to a $JT$-dimensional vector, and $\otimes$ denotes the Kronecker product. The property of rewriting the general covariance  matrix $\Psi \in \mathbb{R}^{JT \times TJ}$ as $\Psi=\Phi \otimes \Sigma$ is called separability condition. Then, the  mean and the variance of the matrix-normal distribution
are:
\begin{align}
\label{props}
    \mathbb{E}(\text{vec}(Z)|M,\Phi,\Sigma)=\text{vec}(M)\quad \text{ and } \quad
    \mathbb{V}(\text{vec}(Z)|M,\Phi,\Sigma)= \Psi.
\end{align}
Being a special case of the multivariate normal distribution, the matrix-normal distribution shares the same properties, like, for instance, closure under marginalization, conditioning and linear transformations \citep{gupta2000matrix}. The separability condition of the covariance matrix has two advantages. First, it allows the modeling of the temporal pattern of interest directly on the covariance matrix $\Phi$. Second, it represents a more parsimonious solution than that of the unrestricted $\Psi$.\\
However, the separability assumption can also be restrictive, as the Kronecker product structure may fail to capture more complex dependencies between rows and columns \citep{dutilleul1999mle, srivastava2008models}. While several procedures exist to formally test separability in multivariate and spatio-temporal data \citep{lu2005tests, mitchell2006kronecker, Pocuca2023Jun}, in our setting it is not a testable hypothesis but rather a structural variable of the proposed model, adopted for its parsimony and interpretability.\\
Moreover, note that there is an identifiability issue regarding the Kronecker product and the parameters $\Phi$ and $\Sigma$ : if $c$ is a strictly positive constant, then $c^{-1}\Phi \otimes c\Sigma=\Phi \otimes \Sigma$. Various solutions have been proposed to solve this issue, including setting $tr(\Phi) = T$ or $\Phi_{11} = 1$ (\cite{Anderlucci2015Jun,gallaugher2018finite}) or impose $|\Phi|=1$ (\cite{Melnykov2018Sep, Tomarchio2022}). We implement the determinant constraint in our approach.\\ 

Introduced by \cite{Viroli2011Oct}, the pdf of the finite Mixture of Matrix-Normals (MMN) model is given by
\begin{equation*}
f(Z|\boldsymbol{\pi},\boldsymbol{\Theta})=\sum_{k=1}^K \pi_k \mathcal{MN}_{(J \times T)}(Z|M_k,\Phi_k,\Sigma_k),
\end{equation*}
where $K$ is the number of mixture components, $\boldsymbol{\pi}=\{\pi_k\}_{k=1}^{K}$ is the vector of mixing proportions, subject to constraint $\sum_{k=1}^{K}\pi_k=1$ and $\boldsymbol{\Theta}=\{\Theta_k\}_{k=1}^K$ is the set of component-specific parameters with $\Theta_k=\{M_k,\Phi_k,\Sigma_k\}$.\\

Throughout this work, we use the terms ``clusters" and ``mixture components" interchangeably, following the common assumption that each component represents a distinct cluster. We acknowledge that this equivalence may not hold in all contexts \citep{Baudry2010Jan}, particularly when multiple components are needed to capture complex cluster shapes.

\subsection{Paper outline}
\label{sec:idea}
As we aim at develop a model easily understandable and interpretable by practitioners with non-statistical background, we found matrix-variate distributions particularly fit, as shown in \cite{Alaimo2023Jan}. Moreover, as noticed in \cite{Anderlucci2015Jun}, the use of matrix-variate distributions allow to drop the conditional independence assumption, frequently implied in longitudinal latent variable models. Despite the efficacy of matrix-variate distributions, up to now these methods have only been applied to continuous data. We introduce a Mixture for Mixed Matrices (MMM) model. Our model expands the use of matrix-variate mixtures to mixed-type data, by building on the framework proposed by \cite{clustmd} and \cite{Choi2023Jul} in the cross-sectional context.\\

In Sections \ref{sec:model} and \ref{sec:inference} we will detail our model and the MCMC-EM algorithm to perform inference, respectively. In Section \ref{sec:simstudy} some results on synthetic data are presented to assess the performance of the model. Finally, in Section \ref{sec:real} an real-world application concerning stock exchange data during the Covid-19 pandemic period is outlined.\\
Source code and supplementary materials are publicly available at \url{https://github.com/FraAmato/MMM-paper}.

\section{The MMM model}
\label{sec:model}
Let denote by $y_{ijt}$ the observation of the $j$-th ($j=1,\dots,J$) variable for the $i$-th ($i=1,\ldots,N$) unit at time $t$ ($t=1,\ldots,T$), that is: imagine to observe $N$ units and measuring $J$ mixed variables $T$ times throughout the course of the study. The $J$ mixed variables variables consist of $C$ continuous variables, $O$ categorical variables (ordinal, binary, and nominal) and $G$ count variables, such that $C + O + G = J$. We are going to assimilate ordinal, binary and nominal variables together as we will treat them in the same way. \\
Let us reorganize this data in a random-matrix form such that we denote the observed record of the $i$-th subject as $Y_i \in \mathbb{R}^{J\times T}$. $\mathbf{Y}=\{Y_i\}_{i=1}^{N}$ is a sample of $(J\times T)$-variate matrix observations $Y_i \in [\mathbb{R}^{C\times T},\mathbb{N}^{O\times T},\mathbb{N}_{0}^{G\times T}]^{\top}$, $J=C+O+G$. The ordinal, binary and nominal levels are coded by non-negative integers such that each variable $O$ has levels $\{1,\ldots,C_o\} \in \mathbb{N}$. In this work, we will consider zero not included in the set of natural numbers. We will use the notation $\mathbb{N}_{0}$ to indicate $\mathbb{N}\cup\{0\}$.\\
Then, we assume that each variable $y_{ijt}$ is the manifestation of an underlying latent continuous variable $z_{ijt}$. 

\subsection{Modeling continuous variables}
Let the subscript $c$ indicate the generic $c$-th continuous variable. We assume that the observed continuous variables $y_{ict}$ matches exactly the latent variable:

$$y_{ict} = z_{ict}$$

\subsection{Modeling categorical ordinal variables}
\label{sec:ordata}
To map ordinal data, we follow \cite{Amato2024Apr}. Let the subscript $o$ indicate the generic $o$-th categorical variable, and let this variable have $C_o$ levels. Let $\gamma_{o}$ denote a $C_{o} + 1$ -dimensional vector of thresholds that partition the real line for the corresponding $o$-th underlying continuous variable, and let the threshold parameters be constrained such that $-\infty = \gamma_{o,0} \leq \gamma_{o,1} \leq\ldots\leq \gamma_{o,C_o}=\infty$. If the latent $z_{iot}$ is such that $\gamma_{o,c_o-1}<z_{iot}<\gamma_{o,c_o}$ then the observed ordinal response takes value $y_{iot} = c_o$.

Moreover, let define $\mathcal{O}^{O \times T}$ the set of ordinal matrices of size $J\times T$ whose elements takes values in $\{1,\ldots,C_o\}$. 
Each element of $\mathcal{O}^{O \times T}$ is called a response pattern. Let $R$ be the cardinality of $\mathcal{O}^{O \times T}$.
Each response pattern $Y_r \in \mathcal{O}^{O \times T}$ is generated by a portion $\Omega_r$ of the latent space $\mathbb{R}^{O \times T}$ according to thresholds $\boldsymbol{\gamma}:=\{\gamma_o \}_{o=1}^O$.
Let the binary vector $\tilde{Y}_i=(\tilde{Y}_{i1},\ldots,\tilde{Y}_{iR})$ 
be one-hot encoding of $Y_i$ such that if the $r$-th pattern is observed then $\tilde{Y}_{ir}=1$ and any other entry in the vector equals zero.\\

A key point is of course the choice of the thresholds $\boldsymbol{\gamma}=\{\gamma_j \}_{o=1}^O$. Although in some cases the thresholds can be part of the inferential process, as it is the case for the polychoric correlation literature \citep{Olsson1979Dec, Olsson1982Sep},  this approach would cause identifiability problems when combined with cluster-specific parameter estimation in our context, as multiple parameter configurations could yield identical likelihoods. Moreoveor, it would substantially increase the computational burden and parameter space dimensionality, potentially leading to convergence issues and overfitting
This is why thresholds are fixed and not considered as parameters.
There are different ways to do it. We decide to follow \cite{Corneli2020Oct}, where the thresholds are chosen as $\gamma_o = (-\infty,1.5,2.5,\dots ,C_o - 0.5,\infty)$.

\subsection{Modeling categorical nominal variables}
For categorical nominal data with $P$ levels we can consider a one-hot encoding for $P-1$ levels and treat them as binary variables. This approach is most suitable for datasets where categorical variables have reasonably balanced numbers of categories, as variable weighting schemes, while theoretically possible, would complicate the inference without necessarily enhancing the clustering objectives. Binary variables can be considered as a special case of ordinal variables where the number of classes $C_o = 2$. The threshold cutting the underlying continuous variable is set to 0.

\subsection{Modeling count variables}
For count data we consider a matrix-variate Poisson-log normal distribution \citep{Silva2023Apr}. Let the subscript $g$ indicate the generic $g$-th count variable, then we assume that $y_{igt}$ follows a Poisson distribution with parameter $\exp(z_{igt})$, where $z_{igt}$ is a term of the $G\times T$ underlying latent matrix following a matrix normal distribution.

\subsection{Joint model}
So, we can think of $Y_i$ as a block matrix, and conveniently split it between the first $C$ rows, representing the observed continuous variables, followed by $O$ rows representing the categorical variables and the remaining $J - C -O = G$ rows, representing the count variables. Notice that the slicing happens just over rows but not over columns. Then, we can write $Y_i = [Y^{\alpha}_i,Y^{\beta}_i,Y^{\gamma}_i]^{\top}$, where $Y^{\alpha}_i \in \mathbb{R}^{C\times T}$ is the block containing the continuous variables and $Y^{\beta}_i \in \mathbb{N}^{O\times T}$ gathers the categorical ones (that we coded via integers) and the binary ones, and $Y^{\gamma}_i \in \mathbb{N}_{0}^{G\times T}$ is the block containing the count variables.\\

We assume that each observed block of matrix $Y_i$ manifests the corresponding block of the latent matrix $Z_i = [Z^{\alpha}_i,Z^{\beta}_i,Z^{\gamma}_i]^{\top}$, with linkages to the observed matrix $Y_i$ depending on the variable type of each element $y_{ijt}$, as described previously.\\
Then, we assume a mixture of matrix-normal distributions on the latent space. We can consequently write

\begin{equation}
\label{eq:z}
f\left(
 \begin{aligned}
Z^{\alpha}_i \\ Z^{\beta}_i \\ Z^{\gamma}_i 
\end{aligned}
 \Bigg| 
\boldsymbol{\pi},\boldsymbol{\Theta} \right)
= \sum_{k=1}^K \pi_k \, \mathcal{MN}_{(J \times T)} \left(
\begin{pmatrix}
M^{\alpha}_k \\ M^{\beta}_k \\ M^{\gamma}_k
\end{pmatrix}
,\Phi_k,
\begin{pmatrix}
        \Sigma^{\alpha \alpha}_k & \Sigma^{\alpha \beta}_k & \Sigma^{\alpha \gamma}_k\\
        \Sigma^{\beta \alpha}_k & \Sigma^{\beta \beta}_k &  \Sigma^{\beta \gamma}_k\\
        \Sigma^{\gamma \alpha}_k & \Sigma^{\gamma \beta}_k &  \Sigma^{\gamma \gamma}_k
\end{pmatrix}
\right).
\end{equation}

From here, we can derive the joint model. To keep notation coherent, let define with $\tilde{Y}^{\beta}_i$ the one-hot encoding of the categorical part of $Y_i$ as described in Section \ref{sec:ordata}.
In addition to $Z_i$, we introduce a latent binary $K$-dimensional allocation vector that indicate whether the unit $i$ belongs to the $k$-th cluster, $\ell_i=(\ell_{i1},\ldots,\ell_{iK})$, such that $\ell_{ik}=1$ if the $i$-th unit belongs to the $k$-th cluster.\\
Recalling the links each kind of observed variables has with the latent ones, we can express our model through the following distributional assumptions:
\begin{align*}
    &\ell_i \sim \mathcal{M}(1,\boldsymbol{\pi}), \, \boldsymbol{\pi}=(\pi_1,\ldots,\pi_K)\\
    &Z^{\alpha}_i|\ell_{ik}=1 \sim \mathcal{MN}_{(C \times T)}(Z^{\alpha}_i|\Theta^{\alpha}_k), \, \Theta^{\alpha}_k =\{M^{\alpha}_k,\Phi_k,\Sigma^{\alpha}_k\},\\
    &Z^{\beta}_i|Z^{\alpha}_i,\ell_{ik}=1 \sim \mathcal{MN}_{(O \times T)}(Z^{\beta}_i|\Theta^{\beta|\alpha}_k), \, \Theta^{\beta}_k =\{M^{\beta|\alpha}_k,\Phi_k,\Sigma^{\beta|\alpha}_k\},\\
    & Z^{\gamma}_i|,Z^{\alpha}_i,Z^{\beta}_i,\ell_{ik}=1 \sim \mathcal{MN}_{(G \times T)}(Z^{\gamma}_i|\Theta^{\gamma|\alpha,\beta}_k), \, \Theta^{\gamma}_k =\{M^{\gamma|\alpha,\beta}_k,\Phi_k,\Sigma^{\gamma|\alpha,\beta}_k\};\\ &\tilde{Y}^{\beta}_i|Z^{\beta}_i,\ell_{ik}=1 \sim \mathcal{M}(1,\xi_i), \, \xi_i = (\mathbf{1}_{\Omega_1}(Z^{\beta}_i),...,\mathbf{1}_{\Omega_R}(Z^{\beta}_i)), \\
    & Y^{\gamma}_{igt}|Z^{\gamma}_{igt} \sim \mathcal{P}(\exp(Z^{\gamma}_{igt})),
\end{align*}

\noindent where $\mathcal{M}$ indicates the multinomial distribution and $\mathbf{1}_{\Omega_r}(Z^{\beta}_i)$ is the indicator function that equals 1 when the elements in $Z^{\beta}_i$ have values that determine the $r$-th pattern. Hence, when $\tilde{Y}^{\beta}_{ir} = 1$, the vector $\xi_i$ is a vector whose $r$-th element equals 1 and all the others equal 0.\\
Further, to avoid assuming the independence between the different blocks, to link the matrix latent distributions we resort to condition on one block to another by using the properties of matrix-variate normal distribution (\cite{gupta2000matrix}). Thus, 
$\Theta^{\gamma|\alpha,\beta}_k := \{M^{\gamma|\alpha,\beta}_k,\Phi_k,\Sigma^{\gamma|\alpha,\beta}_k\}$, more precisely $M^{\gamma|\alpha,\beta}_k = M^{\gamma}_k + \Sigma^{\gamma \cdot}_{k} \Sigma^{-1,\cdot\cdot}_{k}(Z^{\alpha,\beta}_i - M^{\alpha,\beta}_k)$ and $\Sigma^{\gamma|\alpha,\beta}_k = \Sigma^{\gamma \gamma}_{k} - \Sigma^{\gamma\cdot}_{k}\Sigma^{-1,\cdot\cdot}_{k}\Sigma^{\cdot\gamma}_{k}$, and where $\Theta^{\beta|\alpha}_k := \{M^{\beta|\alpha}_k,\Phi_k,\Sigma^{\beta|\alpha}_k\}$, more precisely $M^{\beta|\alpha}_k = M^{\beta}_k + \Sigma^{\beta\alpha}_{k} \Sigma^{-1,\alpha\alpha}_{k}(Y^{\alpha}_i - M^{\alpha}_k)$ and $\Sigma^{\beta|\alpha}_k = \Sigma^{\beta \beta}_{k} - \Sigma^{\beta\alpha}_{k}\Sigma^{-1,\alpha\alpha}_{k}\Sigma^{\alpha\beta}_{k}$.\\

Lastly, Assuming that the observed value pattern of $\tilde{Y}^{\beta}_i$ is $r$ for sake of notation, we can compose the distribution of each observed mixed matrix as

\begin{align}
    Y_i \sim &\sum_{k=1}^{K} \pi_k \, \mathcal{MN}_{(C \times T)}(Z^{\alpha}_i|\Theta^{\alpha}_k) \cdot \int_{\Omega_r} \mathcal{MN}_{(O \times T)}(Z^{\beta}_i|\Theta^{\beta|\alpha}_k) d Z^{\beta}_i   \nonumber \\
    &\cdot \int_{\mathbb{R}} \prod_{t}^{T}\prod_{g}^{G} \mathcal{P}(y^{\gamma}_{igt}|\exp(z^{\gamma}_{igt})) \cdot \mathcal{MN}_{(G \times T)}(Z^{\gamma}_i|\Theta^{\gamma|\alpha,\beta}_k) d Z^{\gamma}_i \, .
\end{align}

\subsection{Likelihood}
\label{sec:lik}
In the following, $\boldsymbol{Z} := \{Z_i\}_{i=1}^N$ , $\boldsymbol{\ell} := \{\ell_i\}_{i=1}^N$ will indicate the ensembles of $Z_i$ and $\ell_i$ respectively, and $\mathbf{Y} := \{Y_i\}^N_{i = 1}$ be the collection of the observed matrices $Y_i$. Finally, the set of unknown parameters to be estimated is $ \boldsymbol{\Theta} := \{\pi_k,M_k,\Phi_k,\Sigma_k\}_{k=1}^K$ .\\

The joint density of $Y^{\gamma}_i, Z^{\gamma}_i, \tilde{Y}^{\beta}_i,Z^{\beta}_i, Z^{\alpha}_i, \ell_i$ is:
\begin{align*}
f(Y^{\gamma}_i, Z^{\gamma}_i, \tilde{Y}^{\beta}_i,Z^{\beta}_i, Z^{\alpha}_i, \ell_i) = &  f(Y^{\gamma}_i| Z^{\gamma}_i, \tilde{Y}^{\beta}_i,Z^{\beta}_i, Z^{\alpha}_i, \ell_i) \cdot f(Z^{\gamma}_i| \tilde{Y}^{\beta}_i,Z^{\beta}_i, Z^{\alpha}_i, \ell_i) \cdot \\
& f(\tilde{Y}^{\beta}_i|Z^{\beta}_i, Z^{\alpha}_i, \ell_i) \cdot f(Z^{\beta}_i| Z^{\alpha}_i, \ell_i) \cdot f(\ell_i).
\end{align*}

We can therefore write the complete log-likelihood as: 
\begin{align}
\mathcal{L}_C(\boldsymbol{\Theta};\mathbf{Y},\mathbf{Z},\boldsymbol{\ell}) & = \prod_{i=1}^N \prod_{k=1}^K \Big[\pi_k \cdot \Big( \prod_{t}^{T} \prod_{g}^{G} \mathcal{P}(y^{\gamma}_{igt}|\exp(z^{\gamma}_{igt})) \Big) \cdot \mathcal{MN}_{(G \times T)}(Z^{\gamma}_i|\Theta^{\gamma|\alpha,\beta}_k) \cdot \nonumber \\
&\mathcal{MN}_{(O \times T)}(Z^{\beta}_i|\Theta^{\beta|\alpha}_k) \cdot \mathcal{MN}_{(C \times T)}(Z^{\alpha}_i|\Theta^{\alpha}_k) \cdot \prod_{r=1}^R \mathbf{1}_{\Omega_r}(Z^{\beta}_i)^{\tilde{Y}^{\beta}_{ir}}
\Big]^{\ell_{ik}}.
\end{align}

By acknowledging the identity $Y^{\alpha}_i = Z^{\alpha}_i$ and the fact that the last term is non-stochastic, and by using the notation of Equation \ref{eq:z}, we can rewrite the complete log-likelihood as: 

\begin{multline}
    \log \mathcal{L}_C(\boldsymbol{\Theta};\mathbf{Y},\mathbf{Z},\boldsymbol{\ell}) = \sum_{i=1}^N \sum_{k=1}^{K} \ell_{ik} \Biggl[ C + \log(\pi_k) -
    \frac{J}{2}\log(|\Phi_k|)-  \frac{T}{2}\log(|\Sigma_k|)- \\
    \frac{1}{2}tr[\Sigma^{-1}_k(Z_i-M_k)\Phi^{-1}_k(Z_i-M_k)^{\top}] \Biggl],
\label{eq:comp_loglk}
\end{multline}

\noindent where $C$ is a constant with respect to the set of parameters $\boldsymbol{\Theta}$.\\
On the other hand, we can define the observed likelihood as $\mathcal{L}_{O}(\boldsymbol{\Theta};\boldsymbol{Y})$, that is: 

\begin{align}
\label{eq:obs_like}
\mathcal{L}_{O}(\boldsymbol{\Theta};\boldsymbol{Y}) := \prod_{i = 1}^N & \Big\{\sum_{k=1}^{K} \pi_k \, \mathcal{MN}_{(C \times T)}(Z^{\alpha}_i|\Theta^{\alpha}_k) \cdot \int_{\Omega_r} \mathcal{MN}_{(O \times T)}(Z^{\beta}_i|\Theta^{\beta|\alpha}_k) d Z^{\beta}_i   \nonumber \\
&\cdot \int_{\mathbb{R}} \prod_{t}^{T}\prod_{g}^{G} \mathcal{P}(y^{\gamma}_{igt}|\exp(z^{\gamma}_{igt}))\times \mathcal{MN}_{(G \times T)}(Z^{\gamma}_i|\Theta^{\gamma|\alpha,\beta}_k) d Z^{\gamma}_i \, \Big\}.
\end{align} 

\section{Inference}
\label{sec:inference}
In our model, we are assuming two different latent (unobserved) variables. Therefore, we will use the Expectation-Maximization (EM) algorithm \citep{EM} to infer the MMM model's parameters. The EM algorithm is well-suited for situations involving latent variables or unobserved data, as it allows for the estimation of model parameters despite the incompleteness.

\subsection{EM-algorithm}
\label{sec:em}
The EM algorithm is an iterative algorithm that alternates two steps: the expectation step (E-step) and the maximization step (M-step).
It start from an initialization $\hat{\boldsymbol{\Theta}}^{(0)}$ of the parameters. Then, let denote with the superscript ${(s + 1)}$ the parameters estimated in the current step and with $(s)$ the ones computed in the previous step.\\
For the MMM model, the E-step consists of evaluating $\mathcal{Q}(\boldsymbol{\Theta},\hat{\boldsymbol{\Theta}}^{(s)}) := \mathbb{E}( \log \mathcal{L}_C(\boldsymbol{\Theta};\mathbf{Y},\mathbf{Z},\boldsymbol{\ell})|\hat{\boldsymbol{\Theta}}^{(s)},\mathbf{Y})$, that is the expectation of the complete log-likelihood conditioned on the parameters computed in the previous step and on the observed data. In the M-step the parameters are updated by maximizing the expected log-likelihood found on the E step, that is $\hat{\boldsymbol{\Theta}}^{(s+1)} := \underset{\Theta}{\arg \max} \, \mathcal{Q}(\boldsymbol{\Theta},\hat{\boldsymbol{\Theta}}^{(s)})$. The iteration process is repeated until convergence on the log-likelihood is met.

\subsection{Initialization}
\label{sec:init}
To find the initial values of $\hat{\boldsymbol{\Theta}}^{(0)}$ mentioned in Section \ref{sec:em}, our proposal is the following.
Identity matrices are chosen for the initialization of the covariance matrices $\Phi_k$ and $\Sigma_k$.\\
For the initialization of $M_k$ and $\pi_k$, two solutions are proposed and tested in Section \ref{sec:sub_init}. The first is a Kmeans++ \citep{Arthur2007Jan} initialization, that is performed on the vectorized data. The second is a multiple random initialization: the mean matrices $M_k$ are chosen by uniform sampling $K$ matrices among the $N$ observed data matrices. Since the EM algorithm is not guaranteed to converge toward a global optimum, the algorithm is applied multiple times and the results with the highest log-likelihood is selected. For simulations in Section \ref{sec:sub_init}, 5 random initializations proved to be enough, but a higher number might be needed for more complex settings.\\
Both the initialization techniques are applied on the latent space, meaning that for count data they are applied on the logarithm of the observed data.

\subsection{E-step}
\label{sec:estep}
As previously stated, the E-step consists of evaluating $\mathcal{Q}(\boldsymbol{\Theta},\hat{\boldsymbol{\Theta}}^{(s)}) := \mathbb{E}( \log \mathcal{L}_C(\boldsymbol{\Theta};\mathbf{Y},\mathbf{Z},\boldsymbol{\ell})|\hat{\boldsymbol{\Theta}}^{(s)},\mathbf{Y})$, that is the expectation of the complete log-likelihood conditioned on the parameters computed in the previous step and on the observed data.\\

We can expand Equation \ref{eq:comp_loglk} as: 

\begin{multline}
    \log \mathcal{L}_C(\boldsymbol{\Theta};\mathbf{Y},\mathbf{Z},\boldsymbol{\ell}) = \sum_{i=1}^N \sum_{k=1}^{K} \ell_{ik} \Biggl[C + \log(\pi_k) -\frac{J}{2}\log(|\Phi_k|)- \frac{T}{2}\log(|\Sigma_k|)-\\ \frac{1}{2}tr[\Sigma^{-1}_k Z_i \Phi^{-1}_k Z^{\top}_i - \Sigma^{-1}_k  Z_i \Phi^{-1}_k M^{\top}_k  -
    \Sigma^{-1}_k M_k \Phi^{-1}_k Z^{\top}_i + \Sigma^{-1}_k M_k 
    \Phi^{-1}_k
    M^{\top}_k] \Biggl].
\label{eq:comp_loglk_open}
\end{multline}

Then, from Equation \ref{eq:comp_loglk_open}, it is easy to see that the expected values to be computed are $\mathbb{E}(\ell_{ik}|\hat{\boldsymbol{\Theta}}^{(s)},\mathbf{Y})$, $\mathbb{E}(\ell_{ik} Z_i|\hat{\boldsymbol{\Theta}}^{(s)},\mathbf{Y})$ and $\mathbb{E}(\ell_{ik} Z_i \Phi^{-1(s)}_k Z_i^{\top}|\hat{\boldsymbol{\Theta}}^{(s)},\mathbf{Y})$ or $\mathbb{E}(\ell_{ik} Z_i^{\top} \Sigma^{-1(s)}_k Z_i|\hat{\boldsymbol{\Theta}}^{(s)},\mathbf{Y})$ by the cyclic property of the trace. As we will see in Section \ref{sec:mstep}, we will need both.\\

We will proceed with their computation one by one. First, $\mathbb{E}(\ell_{ik}|\hat{\boldsymbol{\Theta}}^{(s)},\mathbf{Y})$ can be computed according to the Bayes' rule as

\begin{equation}
    \mathbb{E}(\ell_{ik}|\hat{\boldsymbol{\Theta}}^{(s)},\mathbf{Y}) = 
    \frac{q_{ik}}{\sum^K_{h = 1} q_{ih}} =: \hat{\uptau}^{(s+1)}_{ik}
\end{equation}
where
\begin{align*}
     q_{ik} =  \,& \pi_k \, \mathcal{MN}_{(C \times T)}(Z^{\alpha}_i|\Theta^{(s),\alpha}_k) \cdot \int_{\Omega_r} \mathcal{MN}_{(O \times T)}(Z^{\beta}_i|\Theta^{(s),\beta|\alpha}_k) d Z^{\beta}_i \\ & \cdot \int_{\mathbb{R}} \prod_{t}^{T}\prod_{g}^{G} \mathcal{P}(y^{\gamma}_{igt}|\exp(z^{\gamma}_{igt})) \cdot \mathcal{MN}_{(G \times T)}(Z^{\gamma}_i|\Theta^{\gamma|\alpha,\beta}_k) d Z^{\gamma}_i
\end{align*}
\vspace{10pt}

where the first integral can be approximated through a Monte-Carlo approach applied on the vectorized reparametrization of the matrix-variate distribution and the second one can be approximated by using the estimated value for $Z^{\gamma}_i$ presented in the following.\\

For $\mathbb{E}(\ell_{ik} Z_i|\hat{\boldsymbol{\Theta}}^{(s)},\mathbf{Y})$, recalling the block structure of $Z_i$, we can write

\begin{align}
\label{eq:Zi}
&\mathbb{E}(\ell_{ik} Z_i|\hat{\boldsymbol{\Theta}}^{(s)},\mathbf{Y}) = \mathbb{P}(\ell_{ik}=1  |\hat{\boldsymbol{\Theta}}^{(s)},\mathbf{Y}) \, \cdot \mathbb{E}\left(
\begin{bmatrix}
    Z^{\alpha}_i \\ Z^{\beta}_i \\ Z^{\gamma}_i
\end{bmatrix} 
\middle|\ell_{ik} = 1, \hat{\boldsymbol{\Theta}}^{(s)},\mathbf{Y} \right) = \nonumber \\
& = \mathbb{P}(\ell_{ik}=1  |\hat{\boldsymbol{\Theta}}^{(s)},\mathbf{Y}) \, \cdot 
    \begin{bmatrix}
         Y^{\alpha}_i \\ 
         \mathbb{E}(Z^{\beta}_i|M^{\beta|\alpha,(s)}_k,\Phi^{(s)}_k,\Sigma^{\beta|\alpha,(s)}_k)\\
         \mathbb{E}(Z^{\gamma}_i|M^{\gamma|\alpha,\beta, (s)}_k,\Phi^{(s)}_k,\Sigma^{\gamma|\alpha,\beta,(s)}_k)
    \end{bmatrix} 
:= \hat{\uptau}^{(s+1)}_{ik} \cdot
    \begin{bmatrix}
         Y^{\alpha}_i \\ 
         \hat{M}^{\beta,(s+1)}_{ik}\\
         \hat{M}^{\gamma,(s+1)}_{ik}
    \end{bmatrix},   
\end{align} 

where the matrix-variate expectation related to count data can be computed by defining $z^{\gamma}_i \in \mathbb{R}^{GT \times 1}$ as the vectorized version of $Z^{\gamma}_i$ and computing its expectation $\hat{m}^{\gamma,(s+1)}_{ik} := \mathbb{E}(z^{\gamma}_i |\ell_{ik} = 1,\mathbf{Y} ,\hat{\boldsymbol{\Theta}}^{(s)})$ by means of the sampler implemented in the R package \texttt{Rstan}, that is the R interface to the \texttt{Stan} software \citep{stan}.\\
The matrix-variate expectation related to categorical data can be computed by defining $z^{\beta}_i \in \mathbb{R}^{OT \times 1}$ as the vectorized version of $Z^{\beta}_i$ and computing its expectation $\hat{m}^{\beta,(s+1)}_{ik} := \mathbb{E}(z^{\beta}_i |\ell_{ik} = 1,\mathbf{Y} ,\hat{\boldsymbol{\Theta}}^{(s)})$ through the use of a Gibbs sampler to sample from a truncated multivariate normal distribution.\\

Finally, for $\mathbb{E}(\ell_{ik} Z_i \Phi^{-1}_k Z_i^{\top}|\hat{\boldsymbol{\Theta}}^{(s)},\mathbf{Y})$ we have:

\begin{align}
\label{eq:ZiPhi}
&\mathbb{E}(\ell_{ik} Z_i \Phi^{-1}_k Z_i^{\top}|\hat{\boldsymbol{\Theta}}^{(s)},\mathbf{Y}) = \mathbb{P}(\ell_{ik}=1  |\hat{\boldsymbol{\Theta}}^{(s)},\mathbf{Y}) \, \cdot \mathbb{E}(Z_i \Phi^{-1}_k Z_i^{\top}|\ell_{ik} = 1, \hat{\boldsymbol{\Theta}}^{(s)},\mathbf{Y}) = \nonumber \\  
& = \hat{\uptau}^{(s+1)}_{ik} \cdot  
\begin{bmatrix}
    Y^{\alpha}_i \hat{\Phi}^{-1(s)}_k Y_i^{\alpha \top} & Y^{\alpha}_i \hat{\Phi}^{-1(s)}_k \hat{M}^{\beta,\top(s+1)}_{ik} & 
    Y^{\alpha}_i \hat{\Phi}^{-1(s)}_k \hat{M}^{\gamma,\top(s+1)}_{ik} \\ 
    \hat{M}^{\beta,(s+1)}_{ik} \hat{\Phi}^{-1(s)}_k Y_i^{\alpha \top} &  \hat{D}^{(s+1)}_{ik} &
    \hat{M}^{\beta,(s+1)}_{ik} \hat{\Phi}^{-1(s)}_k  \hat{M}^{\gamma,\top (s+1)}_{ik} \\
    \hat{M}^{\gamma,(s+1)}_{ik} \hat{\Phi}^{-1(s)}_k Y_i^{\alpha \top} &  \hat{M}^{\gamma,(s+1)}_{ik} \hat{\Phi}^{-1(s)}_k  \hat{M}^{\beta,\top (s+1)}_{ik} &  \hat{B}^{(s+1)}_{ik}
\end{bmatrix},
\end{align}

where $\hat{D}^{(s+1)}_{ik} := \mathbb{E}(Z^{\beta}_i \Phi^{-1}_k Z_i^{\beta \top}|\ell_{ik} = 1,\hat{\boldsymbol{\Theta}}^{(s)},\mathbf{Y})$ and  $B^{(s+1)}_{ik} := \mathbb{E}(Z^{\gamma}_i\Phi^{-1}_k Z_i^{\gamma \top}|\ell_{ik} = 1,\hat{\boldsymbol{\Theta}}^{(s)},\mathbf{Y}))$.\\

To compute $D^{(s+1)}_{ik}$ we make use of the the elements of $\hat{S}^{\beta,(s+1)}_{ik} := \mathbb{E}(z^{\beta}_i z_i^{\beta \top}|\ell_{ik} = 1,\mathbf{Y} ,\hat{\boldsymbol{\Theta}}^{(s)})$.
The samples generated to calculate the first moment $\hat{m}^{\beta, (s+1)}_{ik}$ can be reused to compute the matrix $\hat{S}_{ik}^{(s+1)}$ by calculating the mean of the inner product between them.\\
Similarly, for $\hat{B}^{(s+1)}_{ik}$, we make use of the the elements of $\hat{S}^{\gamma,(s+1)}_{ik} := \mathbb{E}(z^{\gamma}_i z_i^{\gamma \top}|\ell_{ik} = 1,\mathbf{Y} ,\hat{\boldsymbol{\Theta}}^{(s)}).$
As before, the samples generated to calculate the first moment $\hat{m}_{ik}^{\gamma,(s+1)}$ can be reused to compute the matrix $\hat{S}_{ik}^{\gamma,(s+1)}$.\\

On the other hand, to compute $\mathbb{E}(\ell_{ik} Z_i^{\top} \Sigma^{-1}_k Z_i|\hat{\boldsymbol{\Theta}}^{(s)},\mathbf{Y})$:

\begin{align}
\label{eq:ZiSigma}
&\mathbb{E}(\ell_{ik} Z_i^{\top} \Sigma^{-1}_k Z_i|\hat{\boldsymbol{\Theta}}^{(s)},\mathbf{Y}) =  \mathbb{P}(\ell_{ik}=1  |\hat{\boldsymbol{\Theta}}^{(s)},\mathbf{Y}) \cdot \mathbb{E}(Z_i^{\top} \Sigma^{-1}_k Z_i|\ell_{ik} = 1, \hat{\boldsymbol{\Theta}}^{(s)},\mathbf{Y}) = \nonumber \nonumber \\
& = \hat{\uptau}^{(s+1)}_{ik} \cdot \Big(Y^{\alpha \top}_i \hat{\Sigma}^{-1, \alpha \alpha}_k Y^{\alpha}_i + Y^{\alpha \top} \hat{\Sigma}^{-1,\alpha \beta}_k \hat{M}^{\beta,(s+1)}_{ik} + Y^{\alpha \top} \hat{\Sigma}^{-1,\alpha \gamma}_k \hat{M}^{\gamma,(s+1)}_{ik} + \nonumber \\ 
& \quad \quad \quad \quad \;\; \hat{M}^{\beta,(s+1) \top}_{ik} \hat{\Sigma}^{-1,\beta \alpha}_k Y^{\alpha}_i + \hat{C}^{(s+1)}_{ik} + \hat{M}^{\beta,(s+1) \top}_{ik} \hat{\Sigma}^{-1,\beta \gamma}_k \hat{M}^{\gamma,(s+1)}_{ik} + \nonumber \\ 
& \quad \quad \quad \quad \;\; \hat{M}^{\gamma,(s+1) \top}_{ik} \hat{\Sigma}^{-1,\gamma \alpha}_k Y^{\alpha}_i + \hat{M}^{\gamma,(s+1) \top}_{ik} \hat{\Sigma}^{-1,\gamma \beta}_k \hat{M}^{\beta,(s+1)}_{ik} + \hat{A}^{(s+1)}_{ik}\Big),
\end{align}
where $\hat{C}_{ik}^{(s+1)} := \mathbb{E}(Z_i^{\beta \top}\Sigma^{\beta \beta}_k Z^{\beta}_i|\ell_{ik} = 1,\hat{\boldsymbol{\Theta}}^{(s)},\mathbf{Y})$, $\hat{A}_{ik}^{(s+1)} := \mathbb{E}(Z_i^{\gamma \top}\Sigma^{\gamma \gamma}_k Z^{\gamma}_i|\ell_{ik} = 1,\hat{\boldsymbol{\Theta}}^{(s)},\mathbf{Y})$ and $\hat{\Sigma}^{-1, **}_k$ indicated the corresponding block of the inverted matrix $\hat{\Sigma}^{-1}_k$ with respect to the notation in Equation \ref{eq:z}. Again, to compute $\hat{C}_{ik}^{(s)}$ we will make use of the elements of $\hat{S}^{\beta, (s+1)}_{ik}$, while for $\hat{A}_{ik}^{(s)}$ we will use the elements of $\hat{S}^{\gamma, (s+1)}_{ik}$.\\

Summing up, this means that computing $\mathbb{E}( \log \mathcal{L}_C(\boldsymbol{\Theta};\mathbf{Y},\mathbf{Z},\boldsymbol{\ell})|\hat{\boldsymbol{\Theta}}^{(s)},\mathbf{Y})$ requires to compute:
\begin{itemize}
    \item $\mathbb{E}(\ell_{ik}|\mathbf{Y},\hat{\boldsymbol{\Theta}}^{(s)}) =: \hat{\uptau}^{(s+1)}_{ik}$,
    \item $\mathbb{E}(z^{\beta}_i|\ell_{ik} = 1,\mathbf{Y},\hat{\boldsymbol{\Theta}}^{(s)}) =: \hat{m}^{\beta, (s+1)}_{ik}$,
    \item $\mathbb{E}(z^{\beta}_i z_i^{\beta \top}|\ell_{ik},\mathbf{Y} ,\hat{\boldsymbol{\Theta}}^{(s)}) =: \hat{S}^{\beta,(s+1)}_{ik}$, whose elements are required for the computation of $\hat{D}^{(s+1)}_{ik}$ and $\hat{C}^{(s+1)}_{ik}$,
    \item $\mathbb{E}(z^{\gamma}_i|\ell_{ik} = 1,\mathbf{Y},\hat{\boldsymbol{\Theta}}^{(s)}) =: \hat{m}^{\gamma, (s+1)}_{ik}$,
    \item $\mathbb{E}(z^{\gamma}_i z_i^{\gamma \top}|\ell_{ik},\mathbf{Y} ,\hat{\boldsymbol{\Theta}}^{(s)}) =: \hat{S}^{\gamma,(s+1)}_{ik}$, whose elements are required for the computation of $\hat{B}^{(s+1)}_{ik}$ and $\hat{A}^{(s+1)}_{ik}$. 
\end{itemize}

\subsection{M-step}
\label{sec:mstep}
The updated for the parameters at step $(s+1)$ are given by

\begin{equation}
    \hat{\pi}^{(s+1)}_k = \frac{\sum_{i=1}^N  \hat{\uptau}^{(s+1)}_{ik}}{N}\quad ,\quad
    \hat{M}^{(s+1)}_k = \frac{1}{\sum_{i=1}^N \hat{\uptau}^{(s+1)}_{ik}} \sum_{i=1}^N \hat{\uptau}^{(s+1)}_{ik} 
    \begin{bmatrix}
        Y^{\alpha}_i \\ \hat{M}^{\beta,(s+1)}_{ik} \\ \hat{M}^{\gamma,(s+1)}_{ik}
    \end{bmatrix}
    ,
\label{eq:pim}
\end{equation}

\begin{align}
   \nonumber \hat{\Sigma}^{(s+1)}_k= \frac{1}{T\sum_{i=1}^N \hat{\uptau}^{(s+1)}_{ik}} & \sum_{i=1}^N \hat{\uptau}^{(s+1)}_{ik} \times \, \\ \nonumber
   &\Bigg(\begin{bmatrix}
    Y^{\alpha}_i \hat{\Phi}^{-1(s)}_k Y_i^{\alpha \top} & Y^{\alpha}_i \hat{\Phi}^{-1(s)}_k \hat{M}^{\beta,\top(s+1)}_{ik} & 
    Y^{\alpha}_i \Phi^{-1(s)}_k \hat{M}^{\gamma,\top(s+1)}_{ik} \\ 
    \hat{M}^{\beta,(s+1)}_{ik} \hat{\Phi}^{-1(s)}_k Y_i^{\alpha \top} &  \hat{D}^{(s+1)}_{ik} &
    \hat{M}^{\beta,(s+1)}_{ik} \hat{\Phi}^{-1(s)}_k  \hat{M}^{\gamma,\top (s+1)}_{ik} \\
    \hat{M}^{\gamma,(s+1)}_{ik} \hat{\Phi}^{-1(s)}_k Y_i^{\alpha \top} &  \hat{M}^{\gamma,(s+1)}_{ik} \hat{\Phi}^{-1(s)}_k  \hat{M}^{\beta,\top (s+1)}_{ik} &  \hat{B}^{(s+1)}_{ik}
\end{bmatrix} - \\ 
& \hat{M}^{(s+1)}_k \hat{\Phi}^{-1(s)}_k \begin{bmatrix}
        Y^{\alpha}_i \\ \hat{M}^{\beta, (s+1)}_{ik} \\ \hat{M}^{\gamma, (s+1)}_{ik}
    \end{bmatrix}^{\top} - \begin{bmatrix}
        Y^{\alpha}_i \\ \hat{M}^{\beta, (s+1)}_{ik} \\ \hat{M}^{\gamma, (s+1)}_{ik}
    \end{bmatrix} \hat{\Phi}^{-1(s)}_k \hat{M}^{\top(s+1)}_k + \hat{M}^{(s+1)}_k \hat{\Phi}^{-1(s)}_k \hat{M}^{\top(s+1)}_k \Bigg)
\end{align}

The update formulas of the two covariance matrices are interconnected:

\begin{align}
\label{eq:m_step_Phi}
\hat{\Phi}^{(s+1)}_k= 
    \frac{1}{J\sum_{i=1}^N \hat{\uptau}^{(s+1)}_{ik}} \sum_{i=1}^N \hat{\uptau}^{(s+1)}_{ik} \Bigg( \nonumber &Y^{\alpha \top}_i \hat{\Sigma}^{-1, \alpha \alpha}_k Y^{\alpha}_i + Y^{\alpha \top} \hat{\Sigma}^{-1,\alpha \beta}_k \hat{M}^{\beta,(s+1)}_{ik} + Y^{\alpha \top} \hat{\Sigma}^{-1, \alpha \gamma}_k \hat{M}^{\gamma,(s+1)}_{ik} + \\ \nonumber
    &  \hat{M}^{\beta,(s+1) \top}_{ik} \hat{\Sigma}^{-1, \beta \alpha}_k Y^{\alpha}_i + \hat{C}^{(s+1)}_{ik} + \hat{M}^{\beta,(s+1) \top}_{ik} \hat{\Sigma}^{-1, \beta \gamma}_k \hat{M}^{\gamma,(s+1)}_{ik} + \\ \nonumber
   & \hat{M}^{\gamma,(s+1) \top}_{ik} \hat{\Sigma}^{-1, \gamma \alpha}_k Y^{\alpha}_i + \hat{M}^{\gamma,(s+1) \top}_{ik} \hat{\Sigma}^{-1, \gamma \beta}_k \hat{M}^{\beta,(s+1)}_{ik} + \hat{A}^{(s+1)}_{ik} - \\ \nonumber 
   &\hat{M}^{\top(s+1)}_k \hat{\Sigma}^{-1(s+1)}_k \begin{bmatrix}
        Y^{\alpha}_i \\ \hat{M}^{\beta,(s+1)}_{ik} \\ \hat{M}^{\gamma,(s+1)}_{ik}
    \end{bmatrix} -\begin{bmatrix}
        Y^{\alpha}_i \\ \hat{M}^{\beta,(s+1)}_{ik} \\ \hat{M}^{\gamma,(s+1)}_{ik}
    \end{bmatrix}^{\top} \hat{\Sigma}^{-1(s+1)}_k \hat{M}^{(s+1)}_k + \\ 
    & \hat{M}^{\top(s+1)}_k \hat{\Sigma}^{-1(s+1)}_k \hat{M}^{(s+1)}_k \Bigg)
\end{align}

At each iteration, after $\hat{\Phi}^{(s+1)}_k$ is computed according to Equation \ref{eq:m_step_Phi}, we will impose $\hat{\Phi}^{(s+1)}_k = \hat{\Phi}^{(s+1)}_k / \,|\hat{\Phi}^{(s+1)}_k|^{1/T}$ to constraint the  determinant of $|\hat{\Phi}^{(s+1)}_k| = 1$, as introduced in section \ref{sec:prel}. Then, $\hat{\Phi}^{(s+1)}_k = (\hat{\Phi}^{(s+1)}_k + \hat{\Phi}^{\top(s+1)}_k) /2$ to enforce symmetry.

\subsection{Convergence}
\label{sec:conv}
Because of the MCMC use during the E-step, the property of monotone increase of the observed log-likelihood does not hold for our model \citep{Ruth2024Jan, McLachlan2007Apr}. Therefore, to asses convergence we use moving average estimation on the observed log-likelihood.\\

Let $l^{s}_o$ the observed log-likelihood at step $s$, then our convergence criterion is 
\begin{equation*}
   \left| \frac{\left( \frac{1}{w_1}\sum^{s}_{i=s-w_1+1} l^{i}_o \right) - \left(\frac{1}{w_2}\sum^{s-w_1}_{i=s-w_1-w_2+1} l^{i}_o\right)}{\frac{1}{w_1}\sum^{s}_{i=s-w_1+1} l^{i}_o} \right| < \varepsilon.
\end{equation*}

In the following, $\varepsilon = 1 \cdot 10^{-3}$ is chosen.

\subsection{Selection of the number of cluster \texorpdfstring{$K$}{K}}
\label{sec:choiceK}
The number of clusters $K$ is selected by minimizing the BIC \citep{Schwarz1978Mar}. We acknowledge that \cite{Tomarchio2025number} found notable performance differences among criteria in matrix-variate mixture contexts, with BIC generally ranking second among the evaluated criteria. Certain criteria showed superior performance in specific configurations, criterion effectiveness varies with data characteristics including dimensionality, sample size, and cluster structure. However, BIC provides consistent and interpretable model selection that aligns with established matrix-variate literature.\\

The BIC for a number of cluster $K$ is defined as
\begin{equation*}
    \text{BIC}_K := -2\log\mathcal{L}_{O}(\boldsymbol{\Theta};\boldsymbol{Y}) + \nu_K \log(N),
\end{equation*}
where $\nu_K$ is the total number of model parameters: 
\begin{equation*}
    \nu_K :=  K[1 + JT + J(J + 1)/2 + T(T + 1)/2]-1,
\end{equation*}
and $\mathcal{L}_{O}(\boldsymbol{\Theta};\boldsymbol{Y})$ is the observed likelihood defined in Equation \ref{eq:obs_like}.\\
To select the model with the optimal number of mixture components $K$, the algorithm is run for K ranging from 1 to $K_{\max}$, where $K_{\max}$is the largest value considered for the number of clusters.

\section{Simulation study}
\label{sec:simstudy}

This section presents numerical experiments on simulated data in order to illustrate the behavior of the proposed model. First, we aim at studying the influence of the initialization procedure and sample size in estimating the partition and the parameters. Secondly, the robustness to different noise ratio in the data concerning the clustering, the parameters estimation and the model selection. Finally, we compare the MMM model to a its continuous counterpart (MMN) when used on mixed data treated like continuous data.

\subsection{Simulation Setup}
\label{sec:sim_set}
A number of 20 different samples have been simulated for increasing number of units $N \in \{100,500,1000\}$, with number of clusters $K = 2$, number of variables $J = 4$, number of times $T = 3$ and cluster proportions $\pi = (0.6,0.4)$. The $J$ variables are of mixed type, with the first variable being continuous, the second being ordinal, the third being binary and the fourth being a counting variable. The ordinal variable has $5$ levels. Each sample has been drawn from a matrix-variate Gaussian and then transformed according to the model described in Section \ref{sec:model}. The distributions parameters were chosen as following: identity matrices for the covariance matrices $\Phi_k$ and $\Sigma_k$ for each cluster, while mean matrices $M_k$ chosen such that the estimated  the optimal Adjusted Rand Index (ARI; \cite{Rand1971Dec}), computed by performing one step of the clustering algorithm using the true parameters, would be around 0.85.  This setting led to the choice of two mean matrices as described in Table \ref{table::simul_means}.\\

Moreover, three scenarios are derived from this setting by adding  some noise by adding to the underlying continuous latent matrix of a percentage $\tau$ of units a reasonable level of noise, generated according to a centered Gaussian with variance equal to 0.5, allocated to the two clusters proportionally to the clusters' size: 0\% (scenario 1), 10\% (scenario 2), 20\% (scenario 3).
The two different kinds of initialization described in Section \ref{sec:init} have been tested.
Regarding the algorithm setup, we set to 100 iterations as the burn-in period of Gibbs sampler in the E-step, and a thinning equal to 2 to prevent too correlated samples. The number of simulated samples is set to 100. 
Concerning the simulation done via \texttt{stan}, we set the chain iterations to 500, of which half as burn-in, for 3 different chains. 

\subsection{Computational time}
Computation time for one iteration on 2.40 GHz 11th Gen Intel Core i5-1135G7 with 16 Go RAM for one step of the algorithm with Kmeans++ initialization for $K = 1$ is about 5 seconds for $N  = 100$ and about 30 seconds second for $N = 1000$. 

\subsection{Influence of initialization \& sample size}
\label{sec:sub_init}

\begin{figure}[ht!]
\centering
\includegraphics[scale = .7]{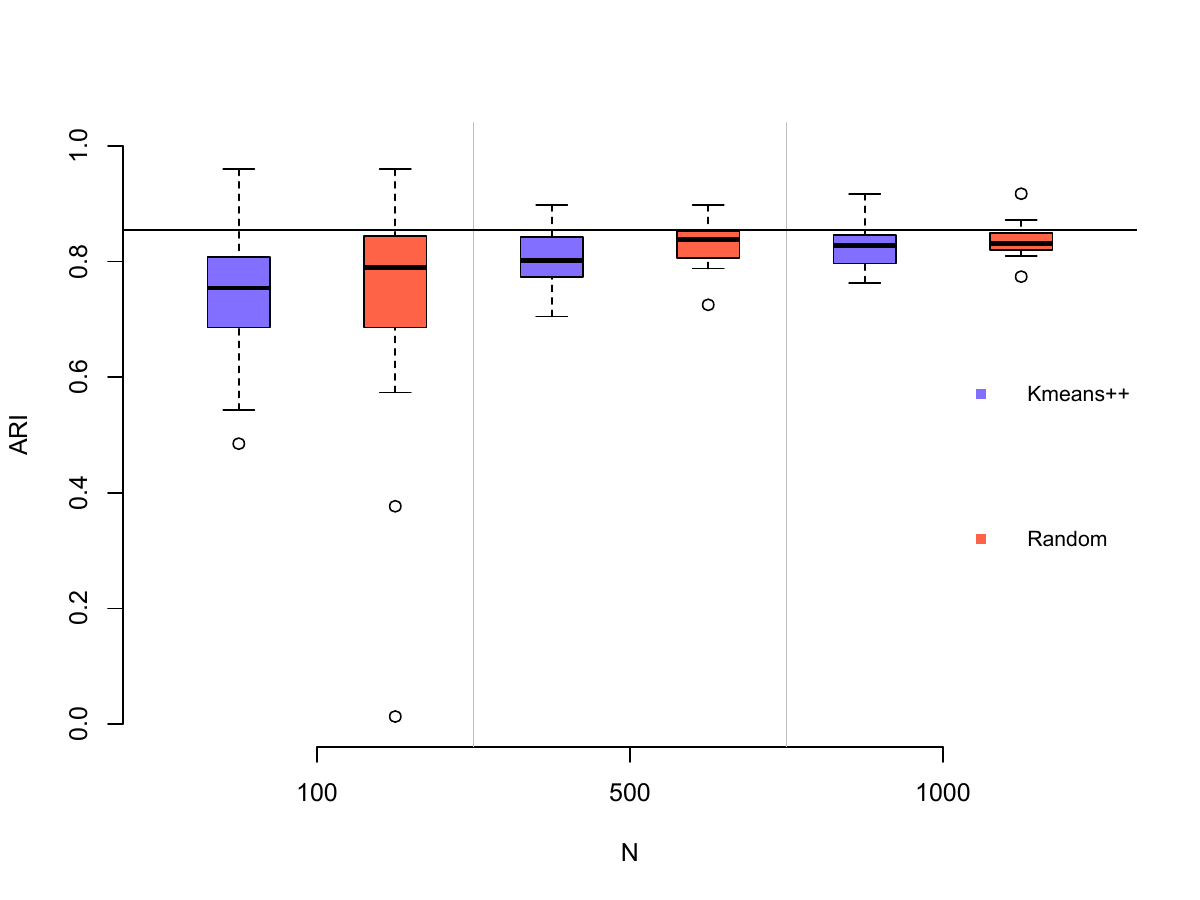}
\caption{Influence of initialization and sample size. The horizontal line represents the estimated Bayesian error.}
\label{fig:inf_init}
\end{figure}

We first aim at studying the ability of the algorithm to recover the simulated model depending on the type of initialization of the EM algorithm and on the size of the sample. Figure \ref{fig:inf_init} shows the quality of estimated partitions assessed by means of ARI. We recall that an ARI of 1 indicates that the partition provided by the algorithm is perfectly aligned with the simulated one. Conversely, an ARI of 0 indicates that the two partitions could as well be some random matches. On the graph, the optimal ARI ($\approx 0.85$) according to the simulation scheme is represented by a horizontal line. 
The boxplots show some small differences in the median values of the ARI measurements between the two initialization methods, with the random multistart initialization performing moderately better than its Kmeans++ counterpart, both in terms of median and of lower variability. When the sample size is sufficiently large, the result that stems from the multistart initialization almost attains the optimal ARI.

However, while the random multistart initialization seems to perform marginally better than the Kmeans++ from a partitioning point of view, it is noteworthy to consider the computational and temporal burden of the former compared to the latter. One might consider whether the tade-off is worthy on a case-by-case base.\\

\begin{figure}[!ht]
\hspace*{-.2in}
\begin{subfigure}{.5\textwidth}
  \centering
  \includegraphics[width=1.1\linewidth]{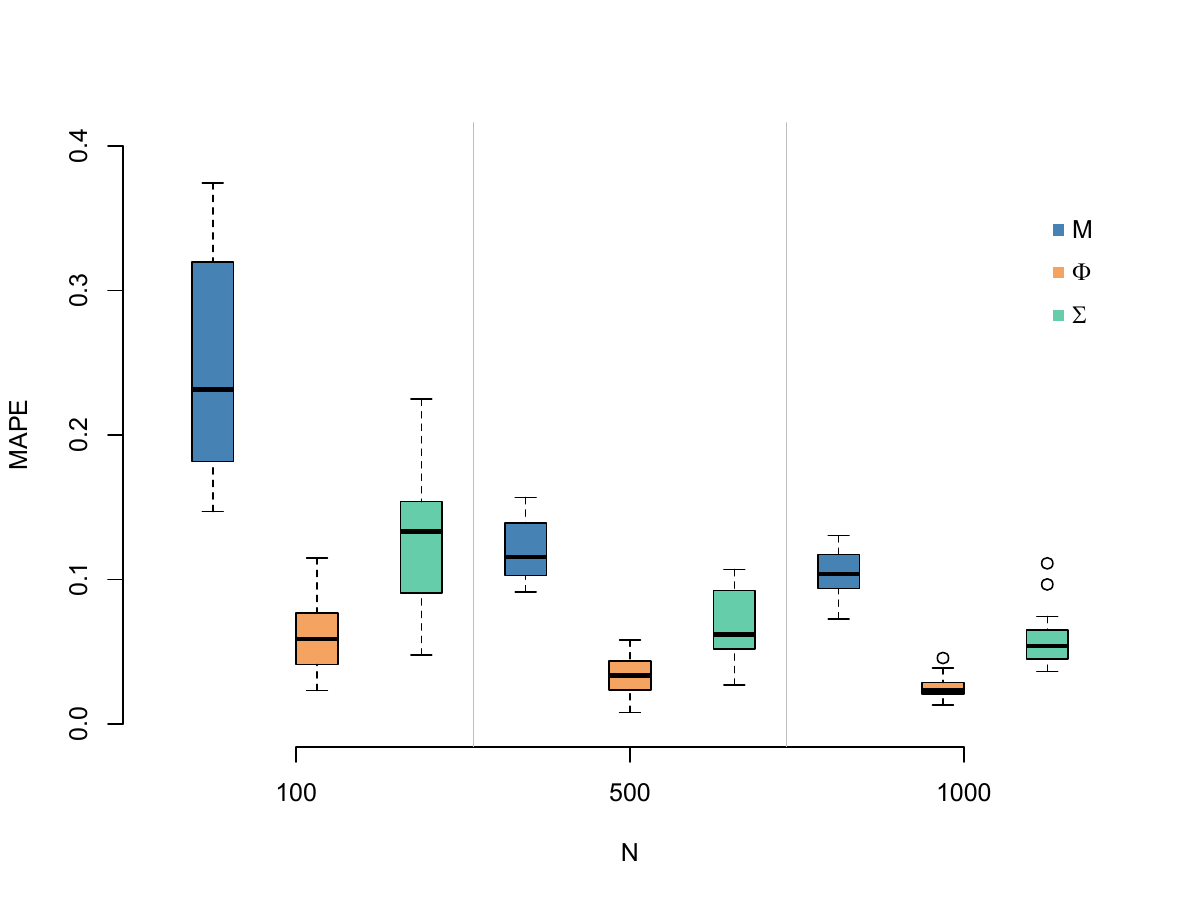}
  \caption{Kmeans++ init}
  \label{fig:mape_kmeans}
\end{subfigure}%
\hspace*{.3in}
\begin{subfigure}{.5\textwidth}
  \centering
  \includegraphics[width=1.1\linewidth]{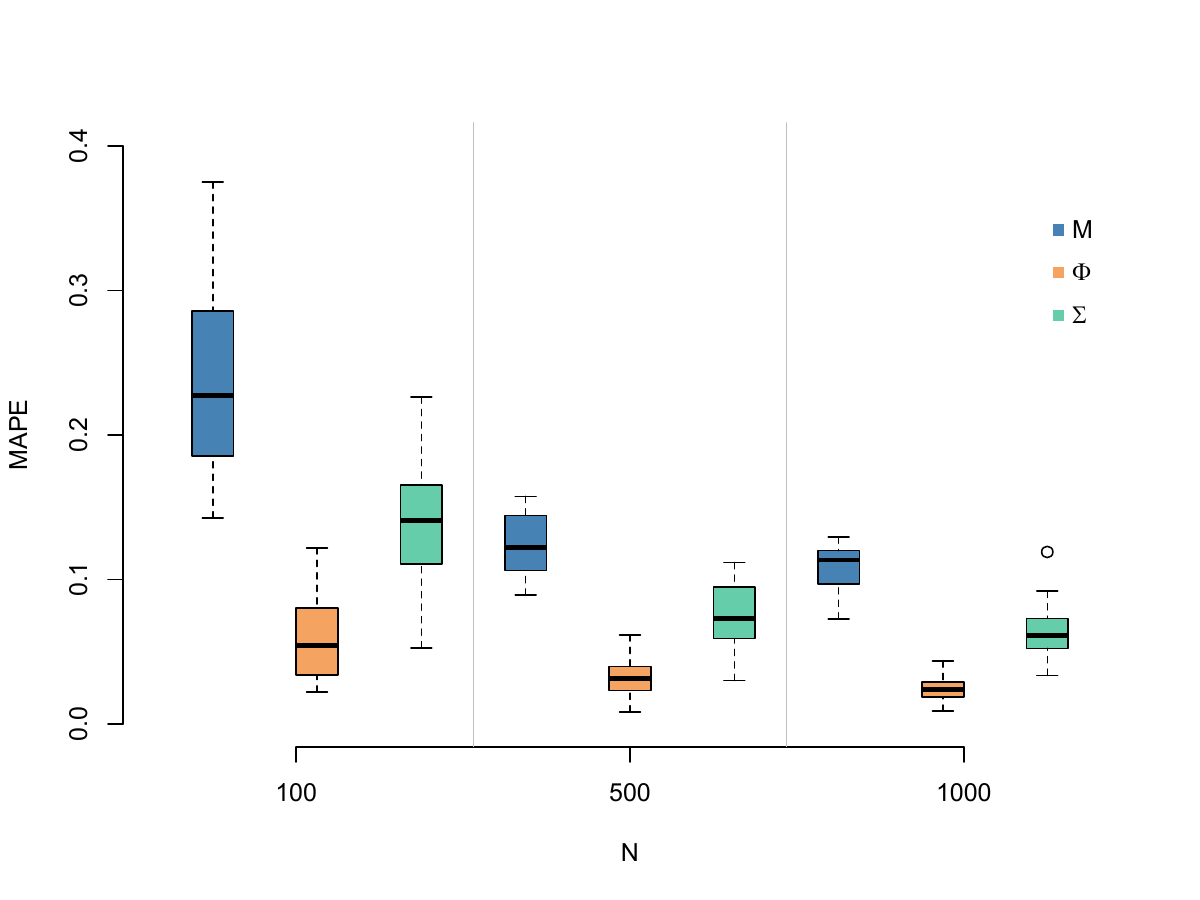}
  \caption{Random init.}
  \label{fig:mape_random}
\end{subfigure}
\hspace*{.5in}
\caption{MAPE for increasing sample size}
\label{fig:MAPE_N_init}
\end{figure}

In addition, we measure their performance also by computing the Mean Absolute Percentage Error (MAPE) on their estimation of the distribution parameters. We recall that the MAPE calculates the average percentage difference between the actual and predicted values of a variable, therefore providing a relative measure of error. For a sample of $N$ units, for a generic parameter $\theta$ it is expressed through the formula:

\begin{equation*}
    \mbox{MAPE}={\frac {100}{N}}\sum _{i=1}^{N}\left|{\frac {\theta_{i}-\hat{\theta}_{i}}{\theta_{i}}}\right|,
\end{equation*}

\noindent where $\hat{\theta}_i$ is the estimated parameter and $\theta_i$ is the true parameter. 
MAPE has some limitations, such as the fact that it cannot be used when actual values are zero or close to zero. This is why for the covariance matrices only the diagonal elements are considered. Results are shown in Figure \ref{fig:MAPE_N_init}.

Regarding the parameters estimates with respect to the different initialization strategies, there is no clear difference in terms of MAPE, with the Kmeans having thinly better rendering.\\

Concerning overall the influence of the sample size, the model behaves as expected: as the sample size increases, the partitioning capabilities improve and tend towards the optimal error. The same happens when we observe the errors concerning the parameter estimations for both the initialization procedures, and the median MAPE values appear to reach a stable value already for $N = 500$, while the values improve further for $N = 1000$, especially in terms of lower variability.\\

Given minimal performance differences, we use Kmeans++ initialization for computational efficiency.\\
Last, while the general magnitude of the MAPE can seem important, it is important to recall that we use a convergence tolerance of $\varepsilon = 1\cdot 10^{-3}$, as per Section \ref{sec:conv}. Better results can be found by reducing the $\varepsilon$, while making the execution more time-consuming.

\subsection{Robustness to noise}

\begin{figure}[ht!]
\centering
\captionsetup{justification=centering}
\includegraphics[width = 1\linewidth]{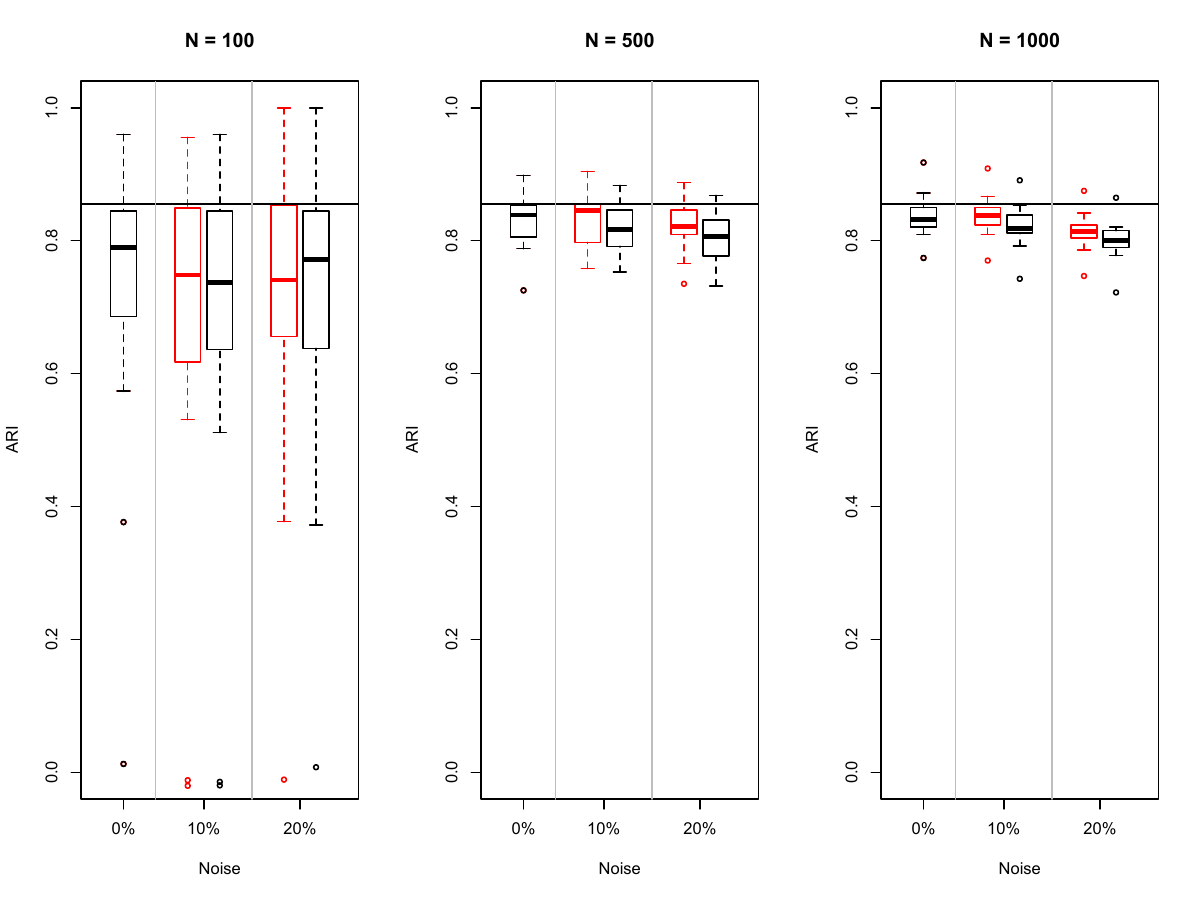}
\caption{ARI for increasing noise proportions and increasing N.\\
In red the ARI for non-noisy units, in black for all of them.}
\label{fig:ARI_noise}
\end{figure}

As written in Section \ref{sec:sim_set}, we also simulated some noisy data to study the robustness of the model in presence of some noise. ARI for different noise proportions were measured and the results are visible in Figure \ref{fig:ARI_noise}. We decided to measure two quantities: the overall ARI for all the units and the ARI just for the non-noisy ones.\\

As we would expect, the overall quality of partitioning estimates slightly decreases as the level of noise increases, indicating that the model is actually disturbed by the noise.\\
Interestingly but somehow to be expected, when $N$ increases the model is more disturbed by the noise, as there are more units affected by it. Moreover, the noise affects the allocation estimation of non-noisy units as well, and again this estimation seems to be more disturbed for a larger $N$.

\subsection{Model selection}
Following the setup described in Section \ref{sec:sim_set}, by varying $N \in \{100,500,1000\}$ and adding increasing noise ratios $\tau \in \{0,0.1,0.2\}$, 9 different scenarios have derived for testing the model selection capabilities. We recall that for each scenario and each $N$, 20 data sets have been drawn. Model selection has been performed through BIC, as described in Section \ref{sec:choiceK}. The results are shown in Table \ref{table:BIC_ord}.

\begin{table}[ht]
\centering
\begin{tabular}{r|rrrr|rrrr|rrrr}
  \hline 
\multicolumn{1}{c|}{$N/K$} &
\multicolumn{4}{c|}{Scenario $\tau = 0$} & \multicolumn{4}{|c|}{Scenario $\tau =0.1$} & 
\multicolumn{4}{|c}{Scenario $\tau = 0.2$}\\
\hline
 & 1 & \textbf{2} &  3  & 4  & 1 & \textbf{2} &  3  & 4  & 1 & \textbf{2} & 3 & 4 \\ 
\hline
100 & \textbf{14} & 6 & 0 & 0  & \textbf{13} & 7 & 0 & 0  & \textbf{12} & 8 &  0  & 0\\ 
500 & 0 & \textbf{19} & 1 & 0 & 0 & \textbf{20} & 0 & 0 & 0 & \textbf{20} & 0 & 0\\
1000 &  0 & \textbf{17} & 3 & 0 & 0 & \textbf{17} & 2 & 1 & 0 & \textbf{18} & 2 & 0\\

\hline
\end{tabular}
\caption{\label{table:BIC_ord} Frequency of selection of each model $K$ by the model through BIC among the 20 simulated data sets, for increasing $N$. The actual value for $K$ is 2. Kmeans++ initialization. In bold the true value for $K$ and the most frequent $K$ detected for each noise ratio and sample size.}
\end{table}

When the sample size increases, the model converges toward the true model. However, as clearly visible in the table, the model tend to underestimate the true number of clusters when the sample size is not sufficiently large, probably due to the insufficient number of units to estimate the parameters from. In addition, \cite{Tomarchio2025number} note that, in mixtures of matrix-variate normal distributions, BIC tends to select a lower number of mixture components as the matrix dimensionality and the number of groups increases.

\subsection{Comparison with continuous conterpart}
\begin{figure}[ht!]
\centering
\includegraphics[scale = .7]{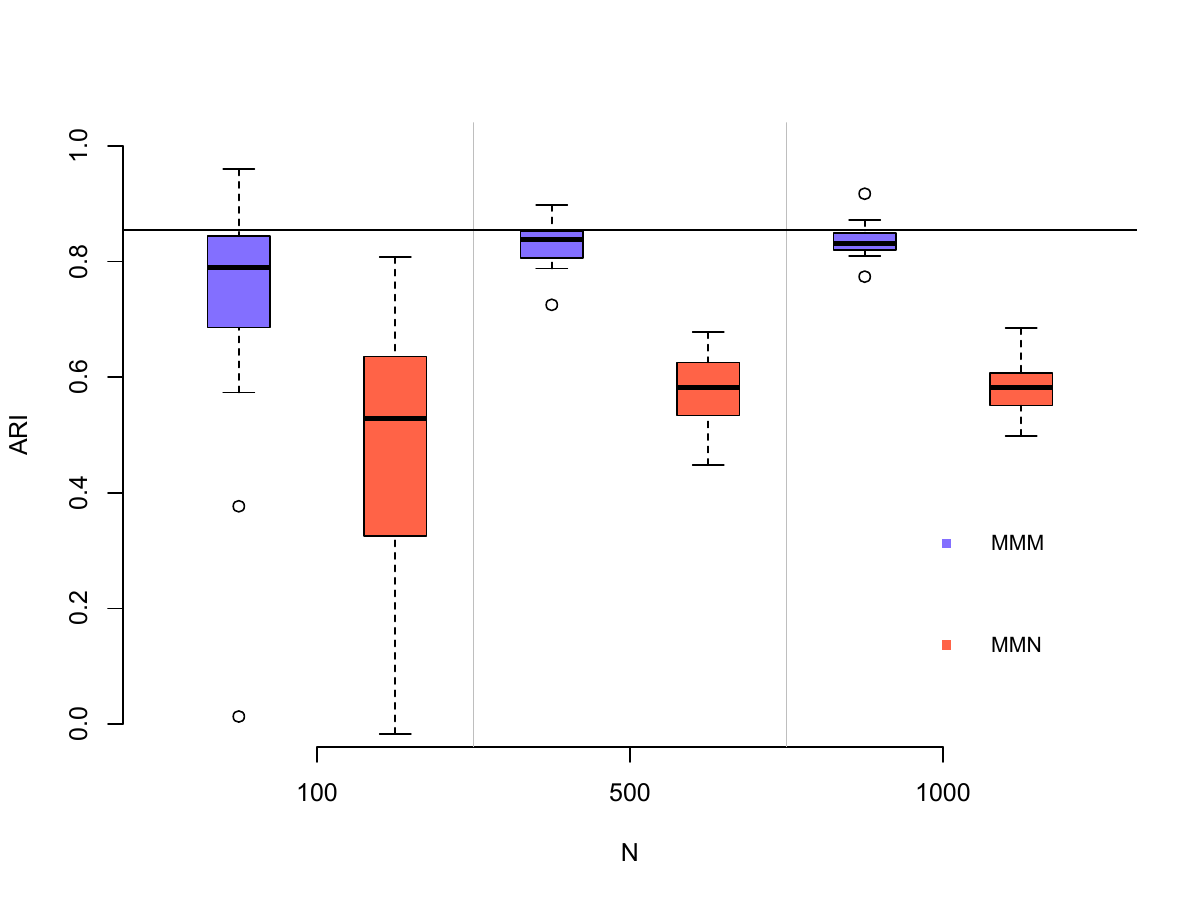}
\caption{Comparison between the MMM and MMN models.}
\label{fig:comp}
\end{figure}

Finally, we compared the MMM model to the classical Mixture of Matrix-Normals (MMN) model, mentioned in Section \ref{sec:related_work}, in a version implemented by us following \cite{Viroli2011Oct}. Essentially, this means treating all the different data-type equally as continuous, as it is often done by practitioners, but keeping the advantages of the matrix-variate structure. The results of the partitioning is presented in Figure \ref{fig:comp}. The hyper-parameters of the competitors have been set to be similar to the one of the MMM in terms of initialization, convergence and covariance matrix parametrization. The MMM model clearly outperforms the MMN model, independently from the sample size.

\begin{figure}[!ht]
\hspace*{-.2in}
\begin{subfigure}{.5\textwidth}
  \centering
  \includegraphics[width=1\linewidth]{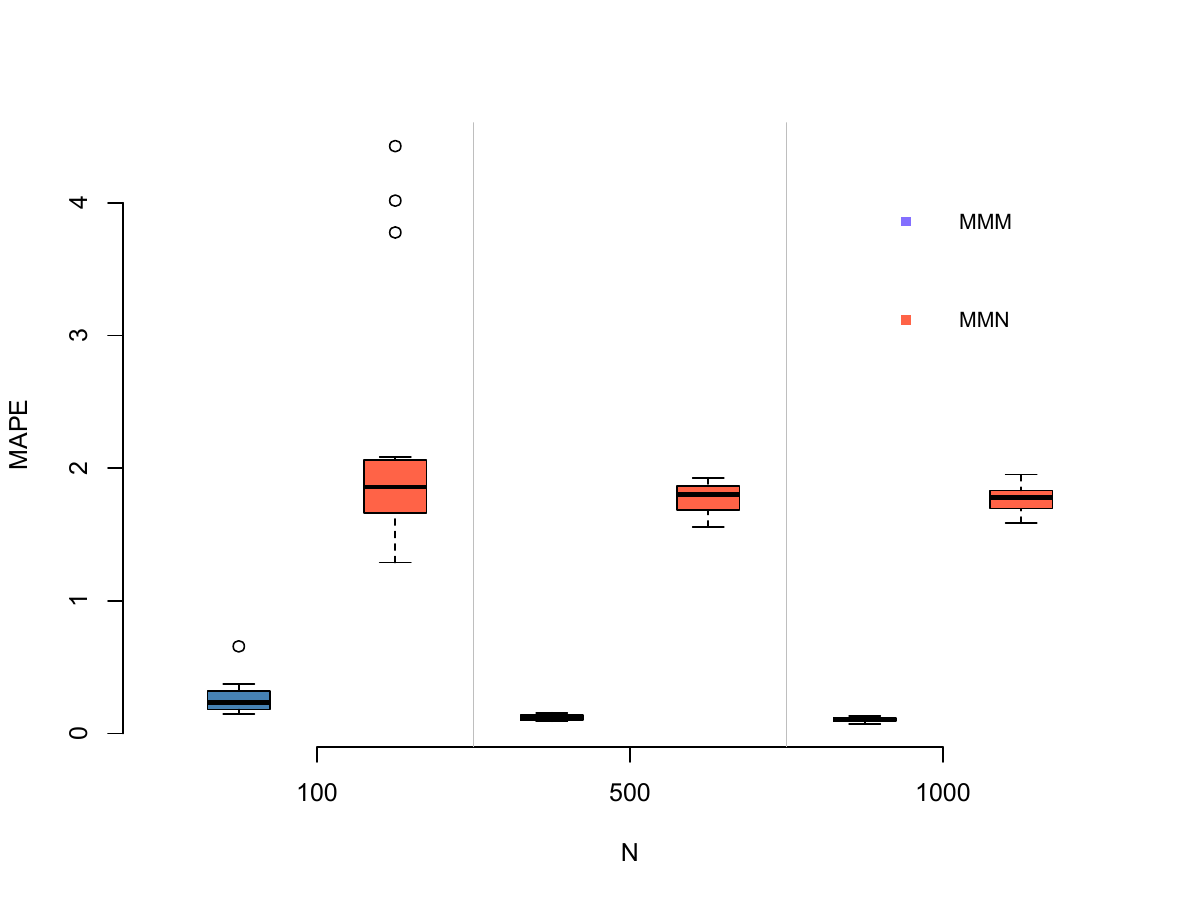}
  \label{fig:mape_M}
  \caption{MAPE for $M$}
\end{subfigure}%
\hspace*{.2in}
\begin{subfigure}{.5\textwidth}
  \centering
  \includegraphics[width=1\linewidth]{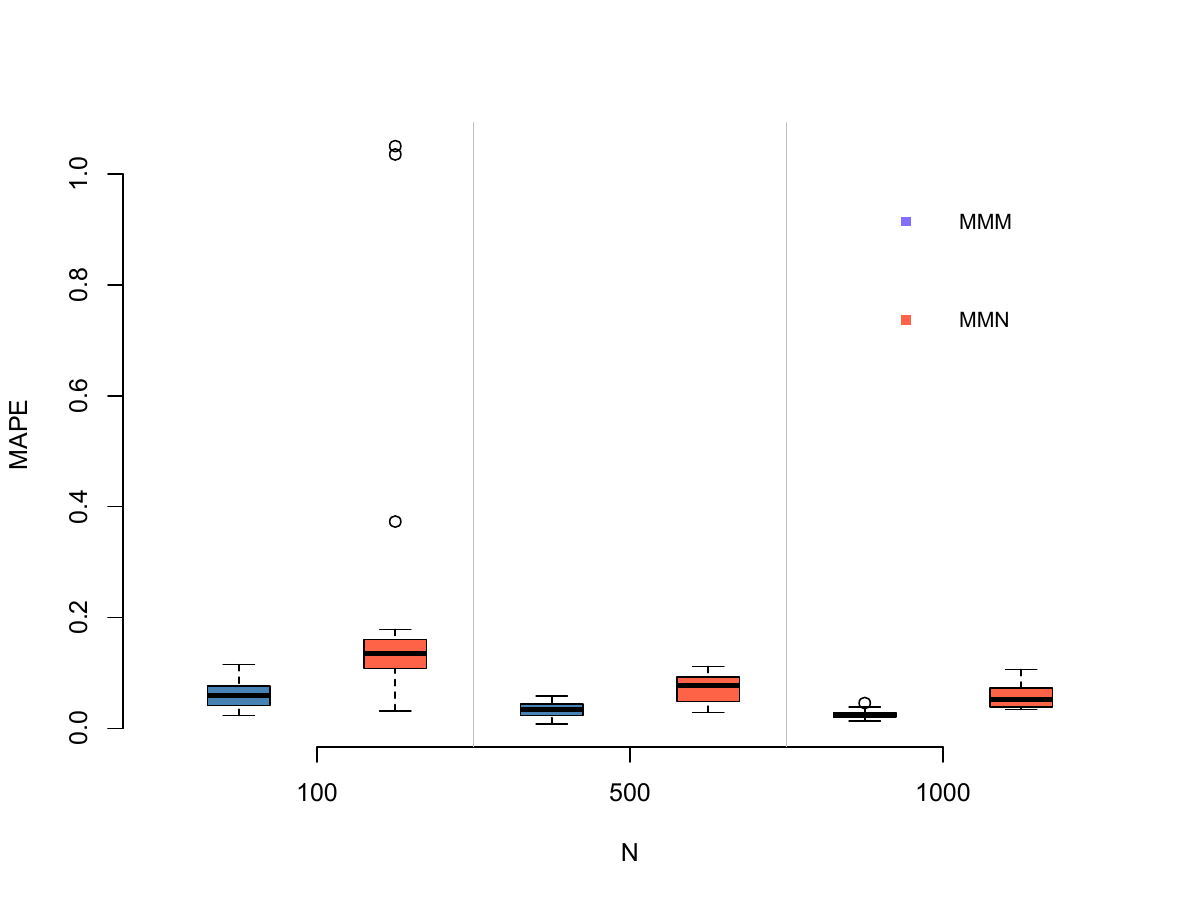}
  \caption{MAPE for $\Phi$}
  \label{fig:mape_Phi}
\end{subfigure}
\hspace*{1.5in}
\label{fig:corMMN}
\begin{subfigure}{.5\textwidth}
  \centering
  \includegraphics[width=1\linewidth]{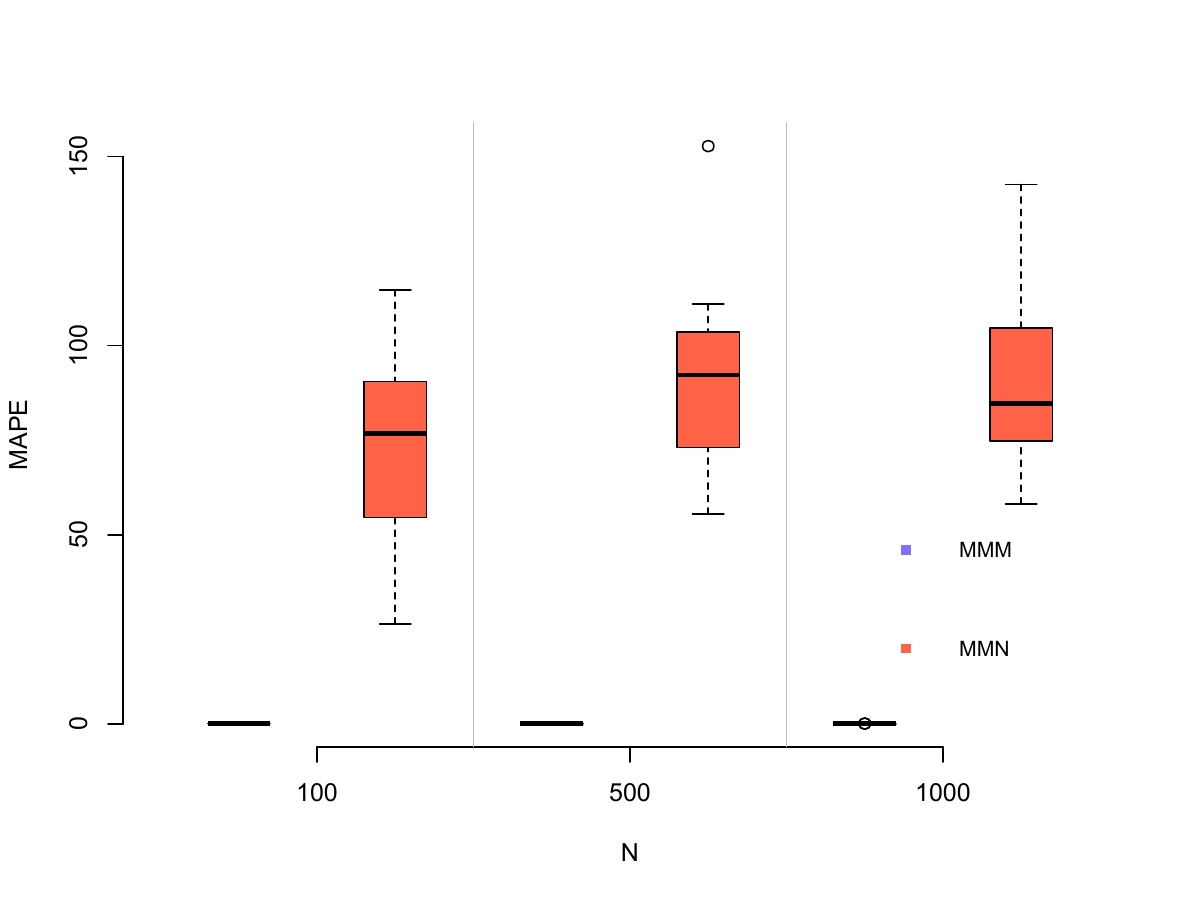}
  \caption{MAPE for $\Sigma$}
  \label{fig:mape_Sigma}
\end{subfigure}
\hspace*{.5in}
\caption{MAPE results for parameter matrices. MMM vs MMN. Kmeans++ init.\\ 
Note the difference in the scales of the MAPE axis in the three graphics.}
\label{fig:MAPE_cont}
\end{figure}

In Figure \ref{fig:MAPE_cont}, we compared the MAPE values for the parameters estimation between MMM and MMN models. The difference between the two is severe, especially for the matrix of means $M$ and the covariance matrix $\Sigma$, while moderate for $\Phi$, due to the constraint on its determinant.The important difference of the results of the MMN model against the MMM one with respect to $M$ and $\Sigma$ is probably due to the count data-type variable. Indeed, without assuming the latent log-normal distribution, the values likely become too out of scale compared to the others.\\

Globally, the experiments described above proved that the MMM model is able to retrieve the true partitioning and to infer the true parameters, even in presence of moderate noise. It is also able select the appropriate number of clusters through BIC when presented with enough sample units. 
We proved that our model outperforms its continuous matrix-variate counterpart when the latter is used to model mixed-type data, as often done by practitioners. We are now confident enough to apply the MMM model on real-world data.

\section{Real-world application}
\label{sec:real}
\subsection{Data description}
The S\&P500 index is a stock exchange index tracking the stock performances of 500 of the largest companies listed on stock exchange market in the United States, where each company is weighted by its market capitalization. It is one of the most commonly followed equity indices and the companies included in the index represent $80\%$ of the total market capitalization of U.S. public companies. While investors are commonly interested in the index in its entirety, it is often the case for them to be interested in the composing companies, reputed as the best ones to invest in, in order to create specific portfolios to be used for long-term investments and wealth management.\\

We collected data concerning companies composing the S\&P500 stock market index. Specifically, we focused on the time period going to the beginning of 2019 to the end of 2023, hence encompassing the period immediately prior and the one immediately succeeding to the  COVID-19 pandemic, which went from the 30th of January 2020 to the 5th of May 2023 according to the World Health Organization (WHO) \citep{Sarker2023May}. The objective of our study is to cluster companies according to their stock behavior during the pandemic period, in order to discover similar patterns during a shock period and possibly adjust our stock portfolio accordingly.\\

For our analysis, we collected for each year and for each listed company the following variables:
\begin{itemize}
    \item \textbf{LogReturns:} continuous variable. The logarithm of the yearly return of the stock. The return is computed as the relative percentage change in the stock adjusted closing price between the first trading day versus the last trading day of the year. In financial analysis, log-returns are often employed instead of the simple returns as log returns have an infinite support (compared to simple returns which are lower-bounded by −100) and as they take into account the compounding effect, making them more suitable for long-term analysis.
    \item \textbf{Grades:} ordinal variable. The investment grade of the stock expressed by institutional investment banks. Specifically, for this study we considered the grades given by ``Bank of America", since it is the institution that releases them for most of the companies of the S\&P500. The grades have three levels: ``Underperform", ``Neutral" and ``Buy". Grades are given multiple times in a year and not all at the same time, so we considered their mode for each fiscal year.
    \item \textbf{Dividends:} binary variable. Whether the stock gave right to a dividend  during the fiscal year or not, regardless of the amount .
    \item \textbf{Volume:} count variable. The total volume of stocks exchanged during the year. Because of the high amount of stocks that are traded during a year, we decided to count per millions of stocks exchanged. Therefore, each counted units will represent a million stocks traded. Generally, securities with higher volume are more liquid. 
\end{itemize}

\noindent Data were collected using the \texttt{pyfinance} Python package.\\
However, grades were not released by Bank of America for all the S\&P500 companies for the entirety of the  time window of our study, but just on 330 of them. We decided to reduce our survey to them. So, overall our dataset is composed of $J = 4$ mixed variables (continuous, ordinal, binary and count) collected for $N = 330$ observations over $T = 5$ time points (years from 2019 to 2023 included). We reorganized these data into a list of matrices.

\subsection{Results}
\begin{figure}[!ht]
\hspace*{-1.5in}
\begin{subfigure}{.5\textwidth}
  \centering
  \includegraphics[width=1\linewidth]{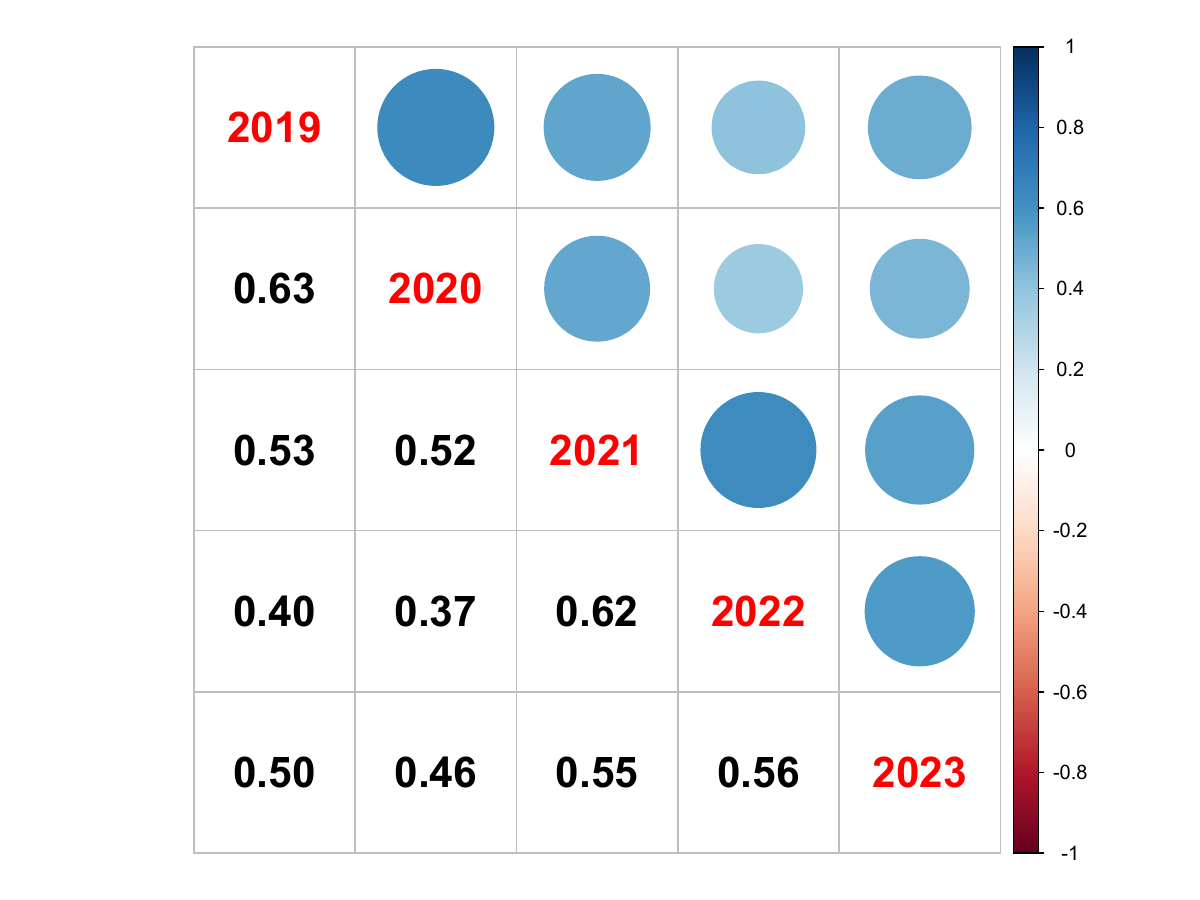}
  \caption{Cluster 1}
  \label{fig:Phi_C1}
\end{subfigure}%
\hspace*{-.8in}
\begin{subfigure}{.5\textwidth}
  \centering
  \includegraphics[width=1\linewidth]{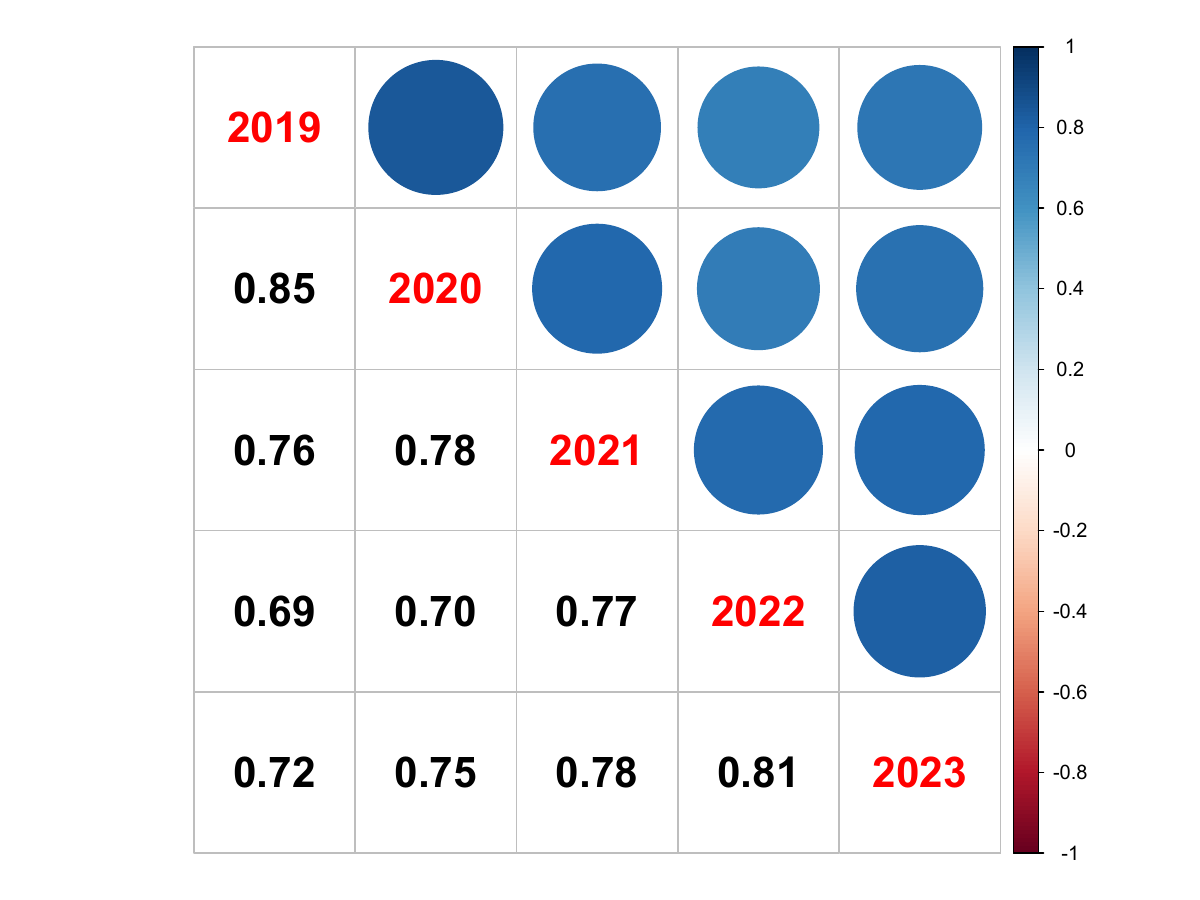}
  \caption{Cluster 2}
  \label{fig:Phi_C2}
\end{subfigure}
\hspace*{-.4in}
\begin{subfigure}{.5\textwidth}
  \centering
  \includegraphics[width=1\linewidth]{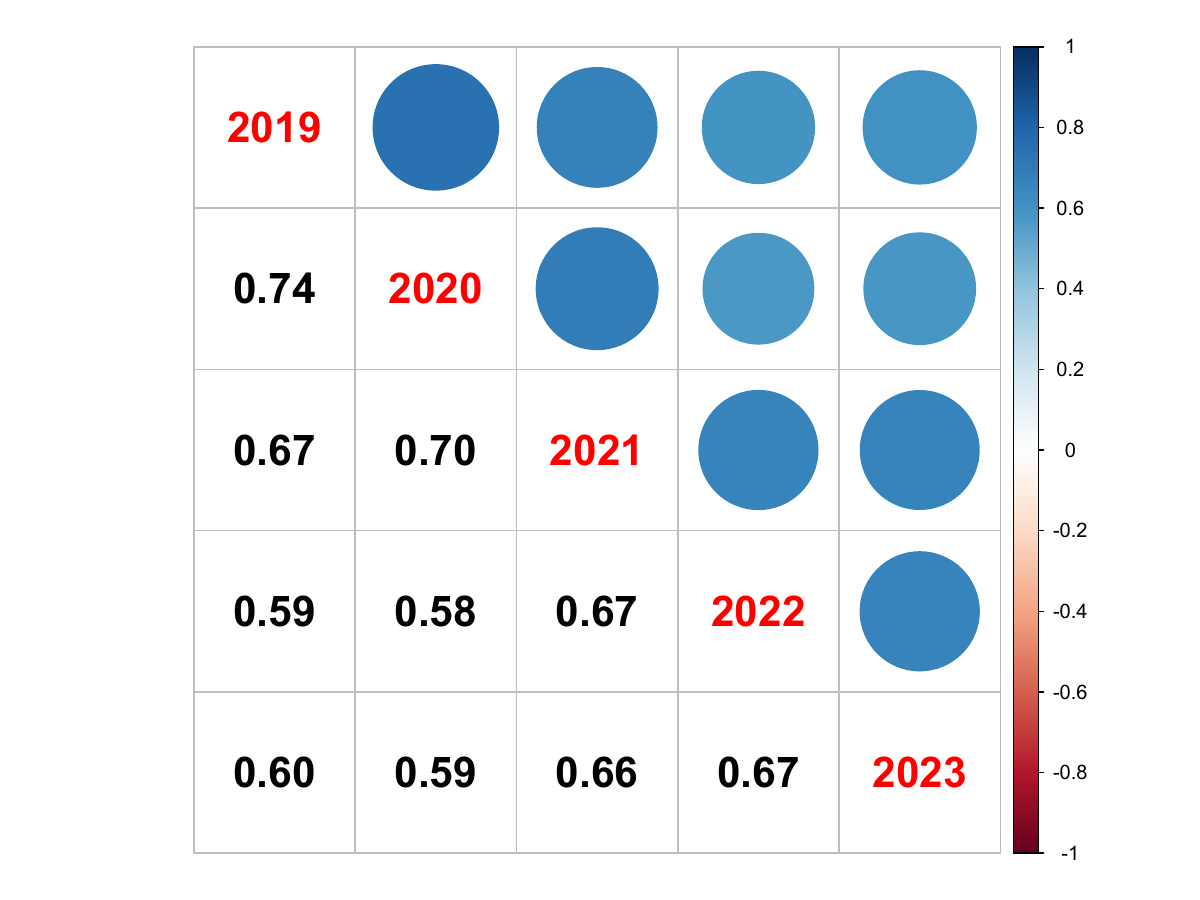}
  \caption{Cluster 3}
  \label{fig:Phi_C3}
\end{subfigure}%
\hspace*{.3in}
\begin{subfigure}{.5\textwidth}
  \centering
  \includegraphics[width=1\linewidth]{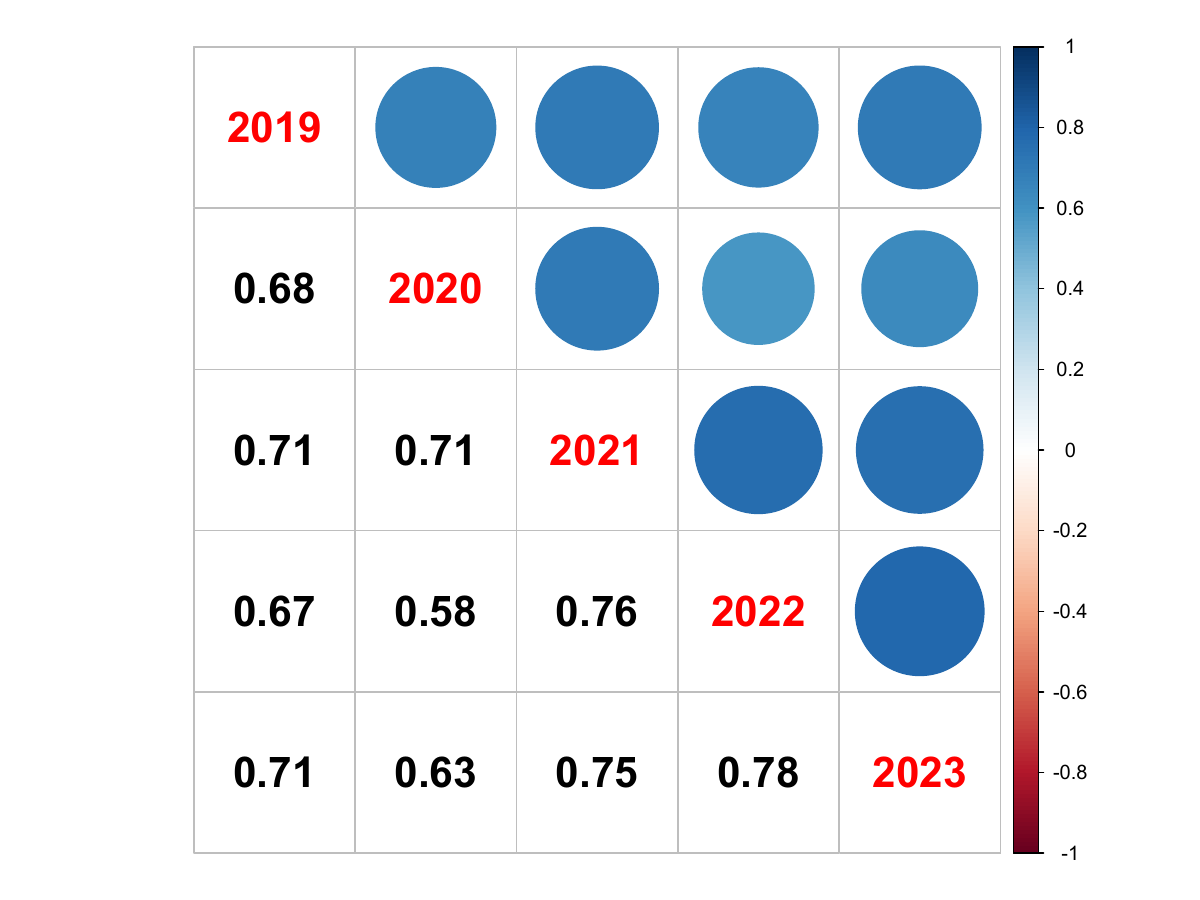}
  \caption{Cluster 2}
  \label{fig:Phi_C4}
\end{subfigure}
\caption{Clusters’ corr-plots among years.}
\label{fig:Phi_corr}
\end{figure}

\begin{figure}[!ht]
\hspace*{-1.5in}
\begin{subfigure}{.5\textwidth}
  \centering
  \includegraphics[width=1\linewidth]{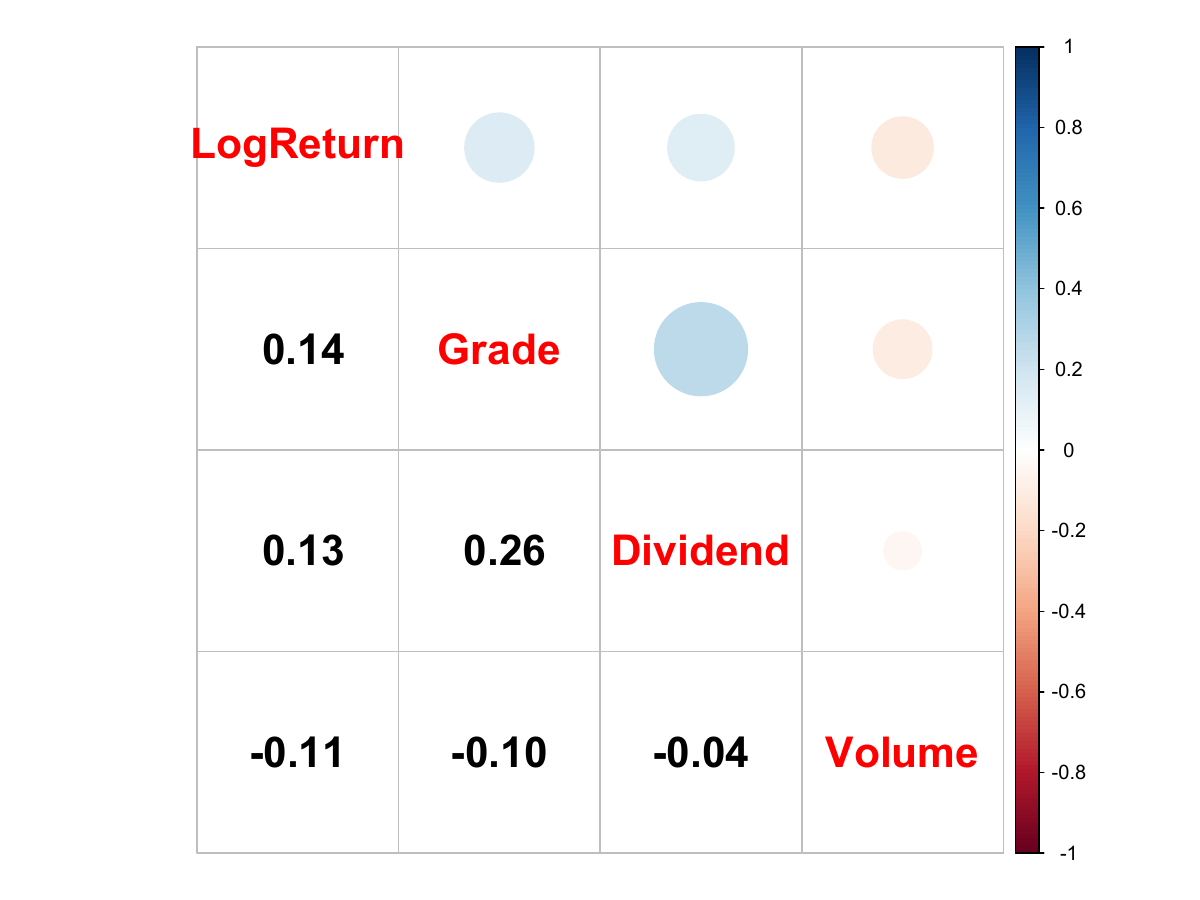}
  \caption{Cluster 1}
  \label{fig:Sigma_C1}
\end{subfigure}%
\hspace*{-.8in}
\begin{subfigure}{.5\textwidth}
  \centering
  \includegraphics[width=1\linewidth]{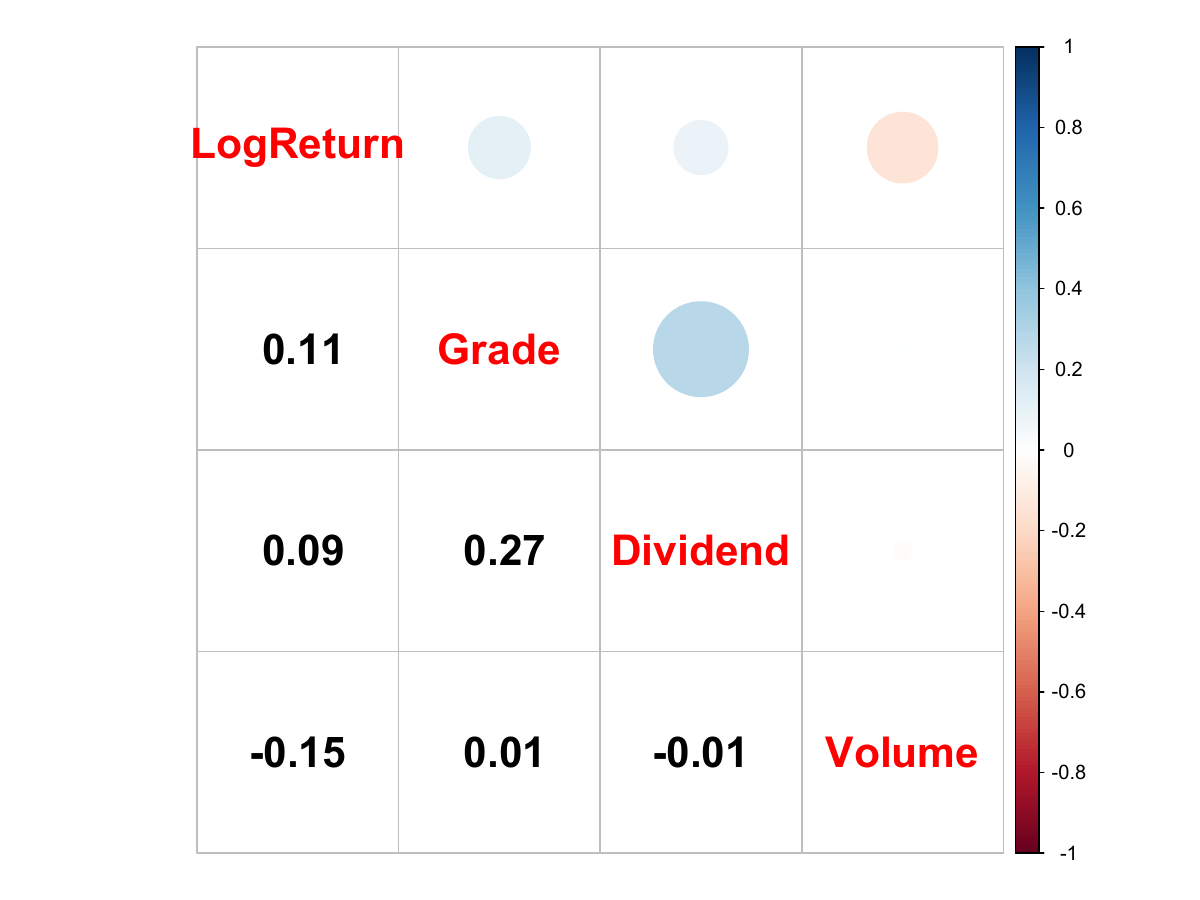}
  \caption{Cluster 2}
  \label{fig:Sigma_C2}
\end{subfigure}
\hspace*{-.4in}
\begin{subfigure}{.5\textwidth}
  \centering
  \includegraphics[width=1\linewidth]{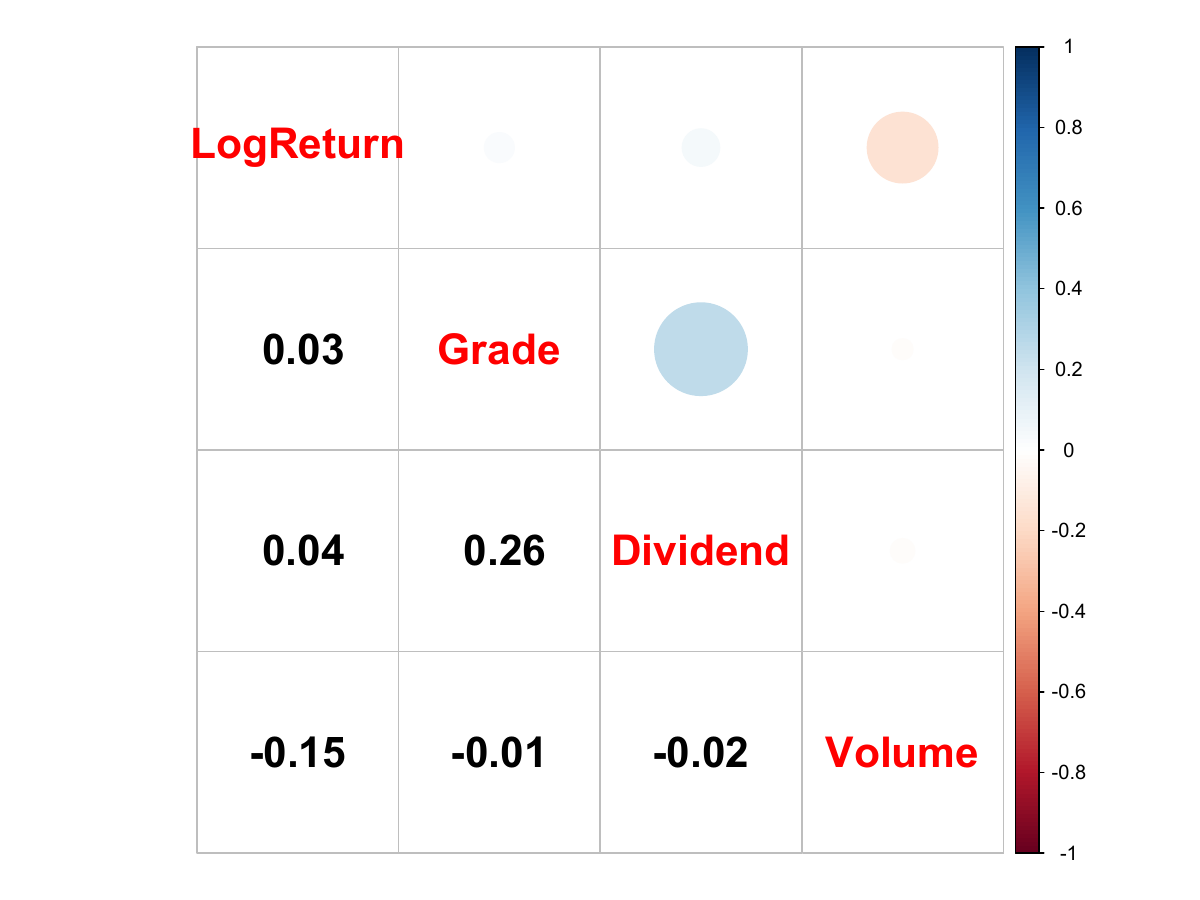}
  \caption{Cluster 3}
  \label{fig:Sigma_C3}
\end{subfigure}%
\hspace*{.2in}
\begin{subfigure}{.5\textwidth}
  \centering
  \includegraphics[width=1\linewidth]{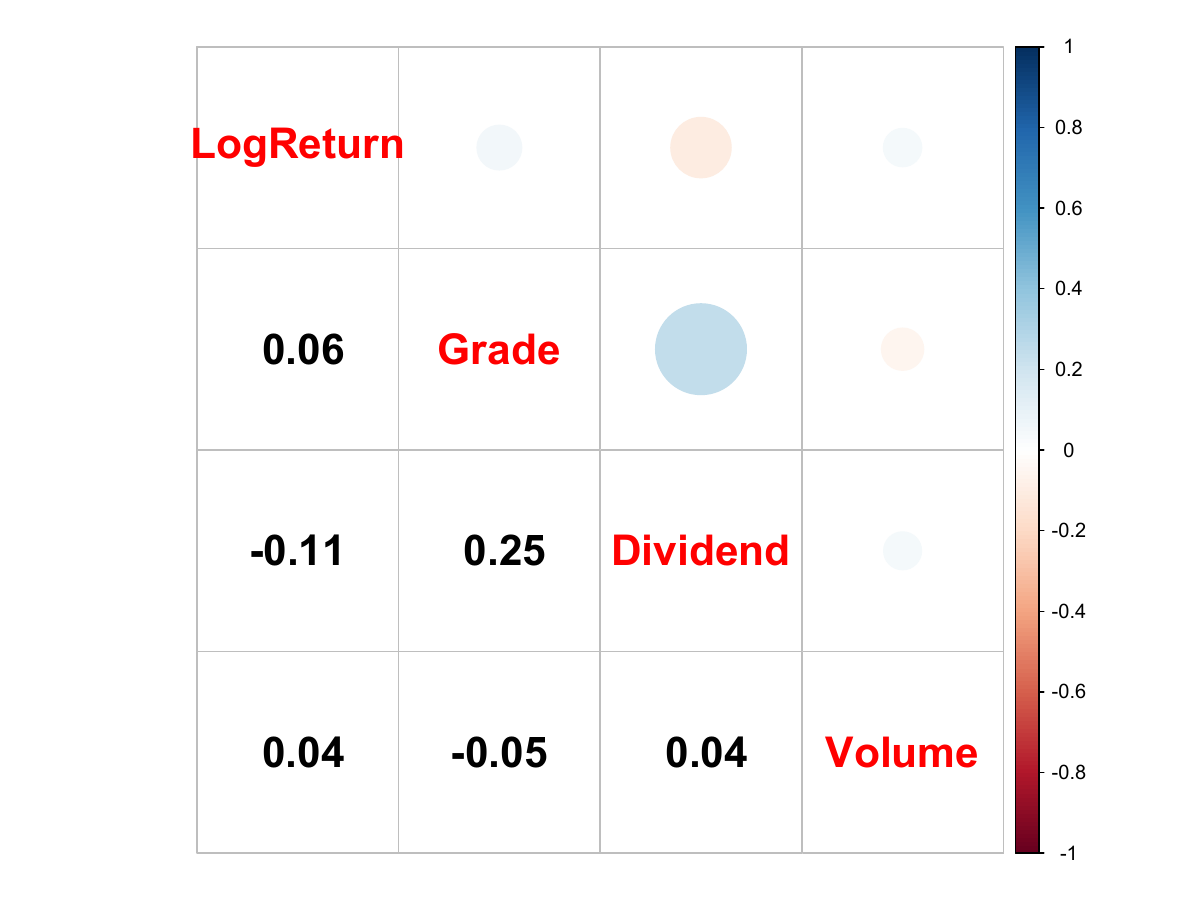}
  \caption{Cluster 4}
  \label{fig:Sigma_C4}
\end{subfigure}
\caption{Clusters’ corr-plots among variables.}
\label{fig:Sigma_corr}
\end{figure}

After performing our clustering algorithm with a number of clusters $K$ ranging from 1 to 8 using Kmeans++ initialization, the model with the lowest BIC is the one with $K$ = 4 (Fig. \ref{fig:BIC_realdata_viz}). The number of units in each cluster is respectively of 94, 50, 154 and 32.\\
The estimated parameters are reported in Table \ref{table:means} for the mean $M$, Table \ref{table:Phi} for the time covariances $\Phi$ and in Table \ref{table:Sigma} for the variable covariances $\Sigma$.
In addition, the correlation matrices are
represented by correlation plots in Figs. \ref{fig:Phi_corr} and \ref{fig:Sigma_corr}, respectively. The tickers of the companies allocated to each cluster is reported at Table \ref{table:stocks}.\\
In Figure \ref{fig:VarMeans} the evolution for the observed outcomes for each cluster is showed.\\
Moreover, by using the ``Global Industry Classification Standard" (GICS) industrial taxonomy developed by ``Standard \& Poor's" (S\&P), we represented the sector composition of each cluster in Figure \ref{fig:cluster_sectors}.\\

\begin{figure}[ht!]
\begin{subfigure}{.5\textwidth}
\centering
\includegraphics[width=.9\linewidth]{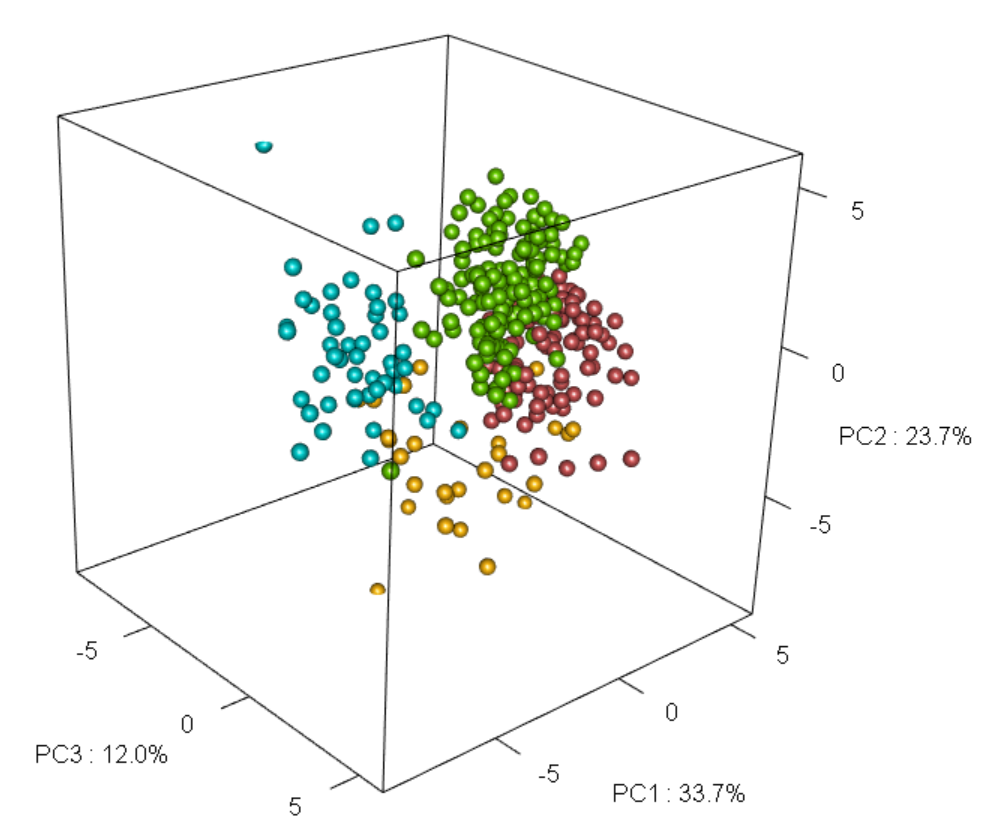}
\caption{Units.}
\label{fig:PCA_units}
\end{subfigure}%
\begin{subfigure}{.5\textwidth}
\centering
\includegraphics[width=.91\linewidth]{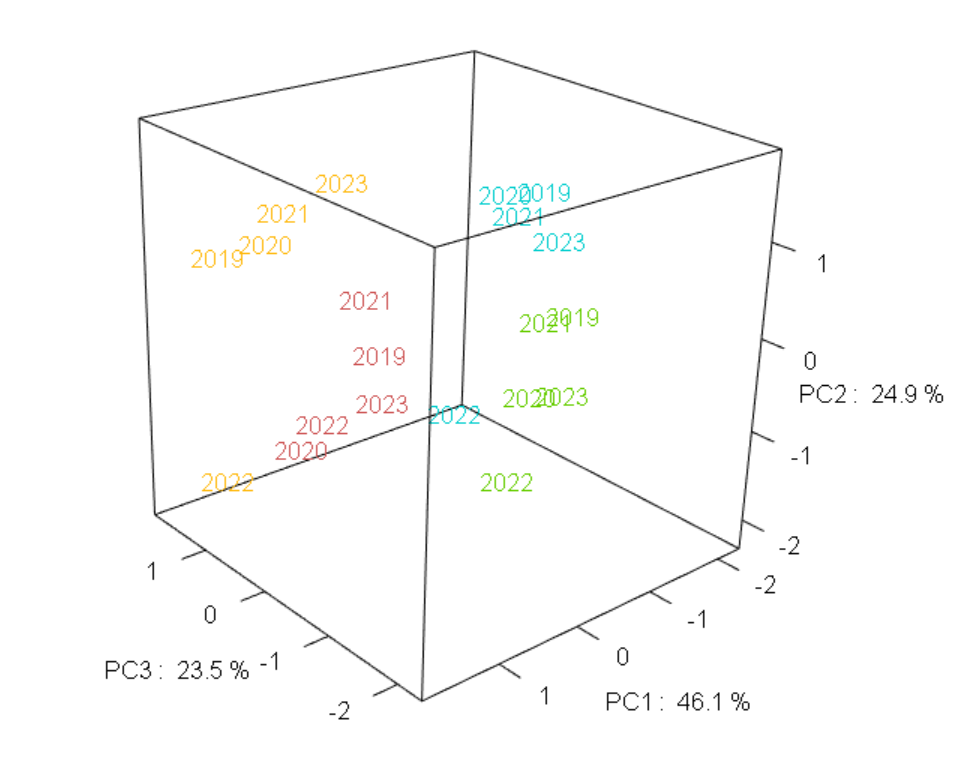}
\vspace{.1cm}
\caption{Cluster means.}
\label{fig:PCA_means}
\end{subfigure}
\hspace*{3cm}
\vspace*{-1.2cm}
\begin{subfigure}{.6\textwidth}
\centering
\includegraphics[width=1\linewidth]{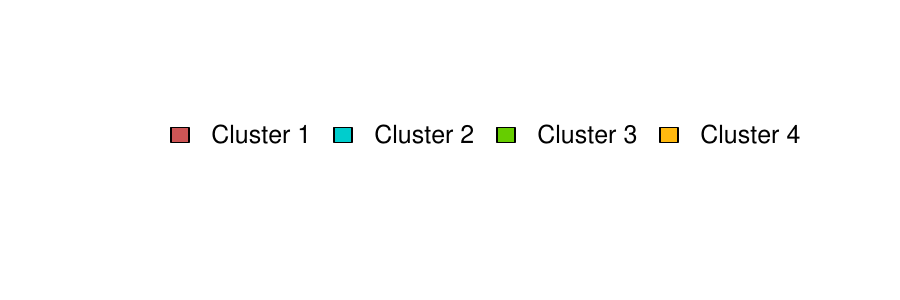}
\end{subfigure}
\caption{Units and cluster means represented through PCA.}
\label{fig:PCA}
\end{figure}

By performing a PCA the latent continuous embedding computed by the MMM model, we can represents the 330 units as in Figure \ref{fig:PCA_units}. A $3D$ representation is provided. For this representation, the temporal structure has been discarded and we have transformed our latent embedding for the units from 4 × 5-dimensional matrices to 20-dimensional vectors. On the other hand, Figure \ref{fig:PCA_means} represents cluster means at each of the 5 years. Such plot allows to visualize the time evolution of each cluster.

\subsection{Interpretation}
\label{sec:interpretation}
First, we can get a preliminary idea by looking at Figure \ref{fig:PCA_units}. As we can see, of the 4 clusters, units belonging to Cluster 1 and Cluster 3 are more concentrated and closer to each other in the latent space, while the ones belonging to Cluster 2 and 4 are more spread out. This is confirmed by looking at Figure \ref{fig:PCA_means}, where the clusters of Cluster 1 and Cluster 3 means occupy adjacent space regions.\\

In the following, we give a summary description for Cluster 4, which we deemed be the most interesting. Interpretations for the other clusters can be found in Appendix \ref{appendix:clust_inter}. 

\begin{figure}[!ht]
\hspace*{-1.5in}
\begin{subfigure}{.5\textwidth}
  \centering
  \includegraphics[width=1\linewidth]{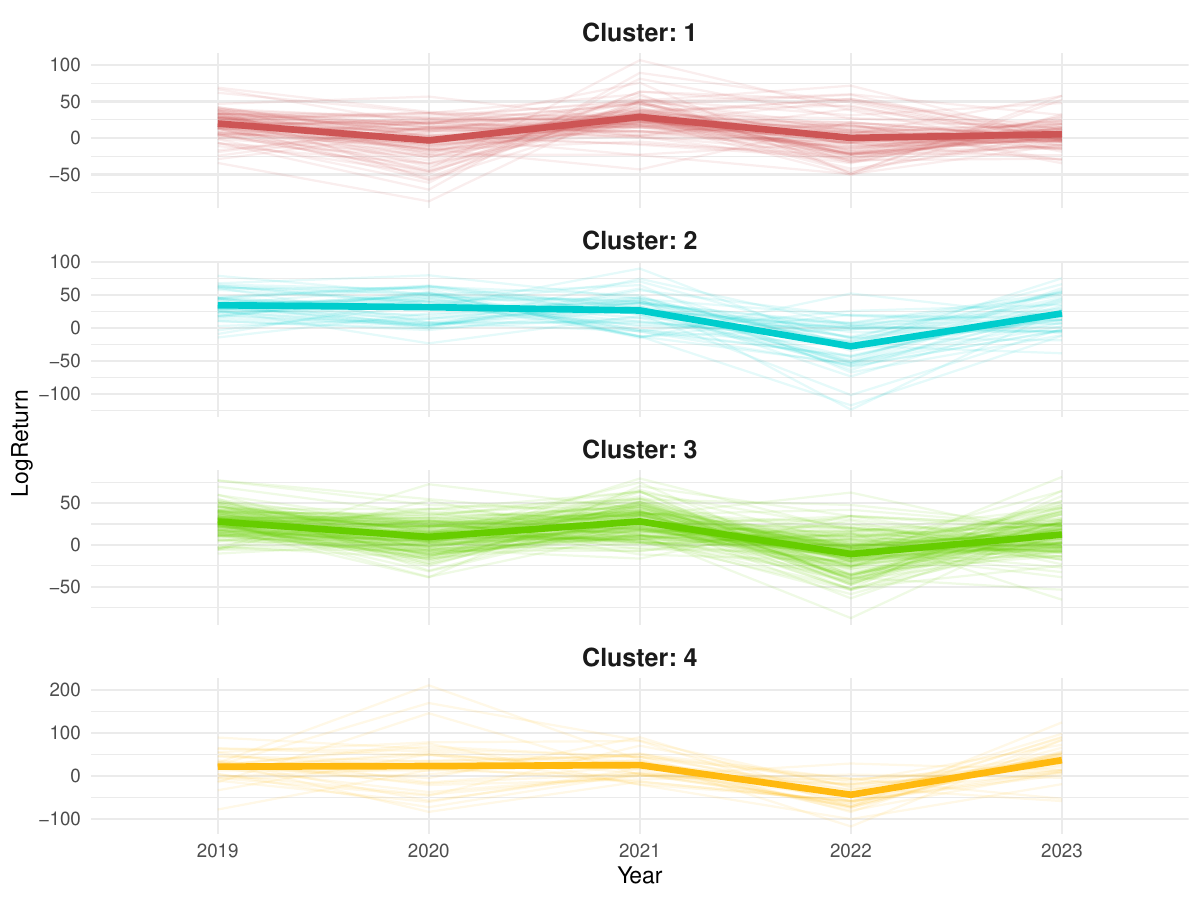}
  \caption{LogReturns}
  \label{fig:LogRet}
\end{subfigure}%
\hspace*{-.8in}
\begin{subfigure}{.5\textwidth}
  \centering
  \includegraphics[width=1\linewidth]{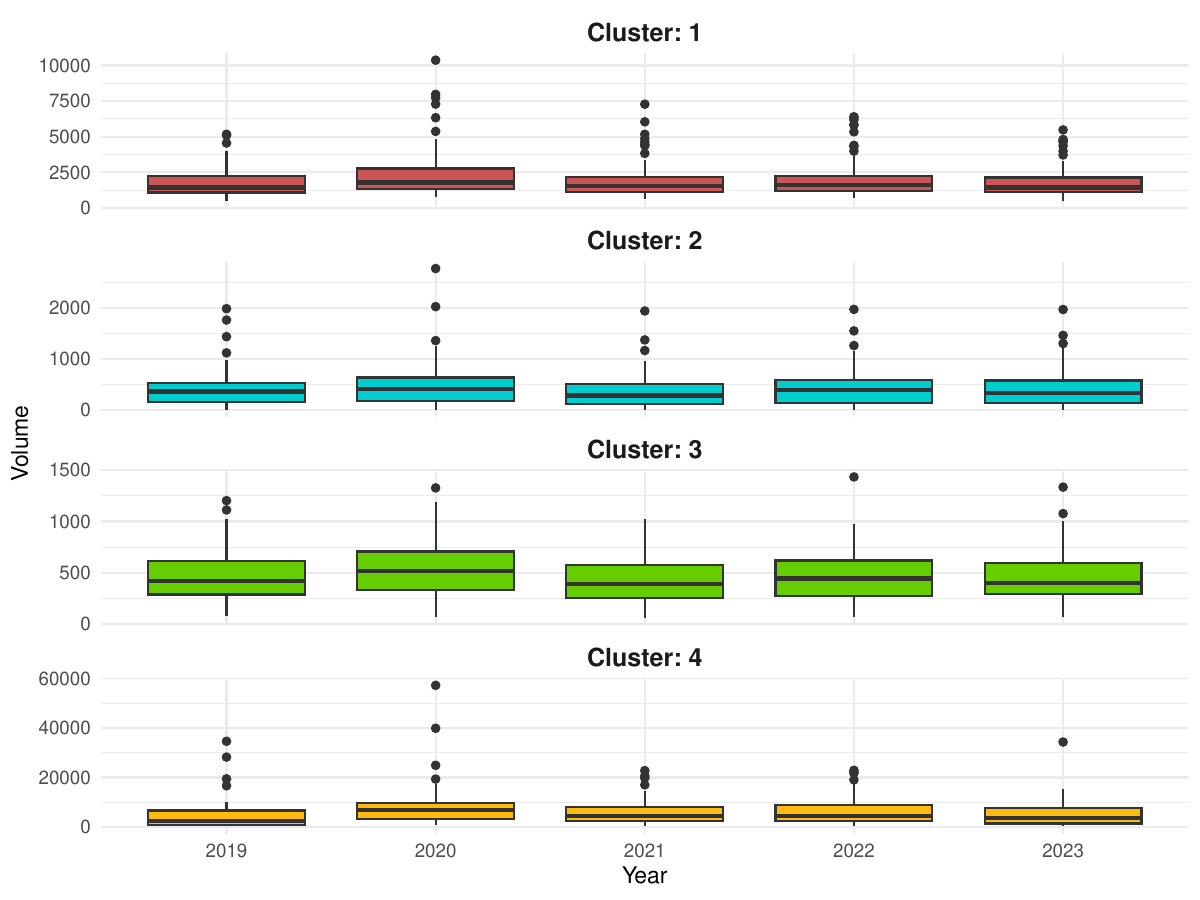}
  \caption{Volume}
  \label{fig:Volumes}
\end{subfigure}
\hspace*{-0.4in}
\begin{subfigure}{.5\textwidth}
  \centering
  \includegraphics[width=1\linewidth]{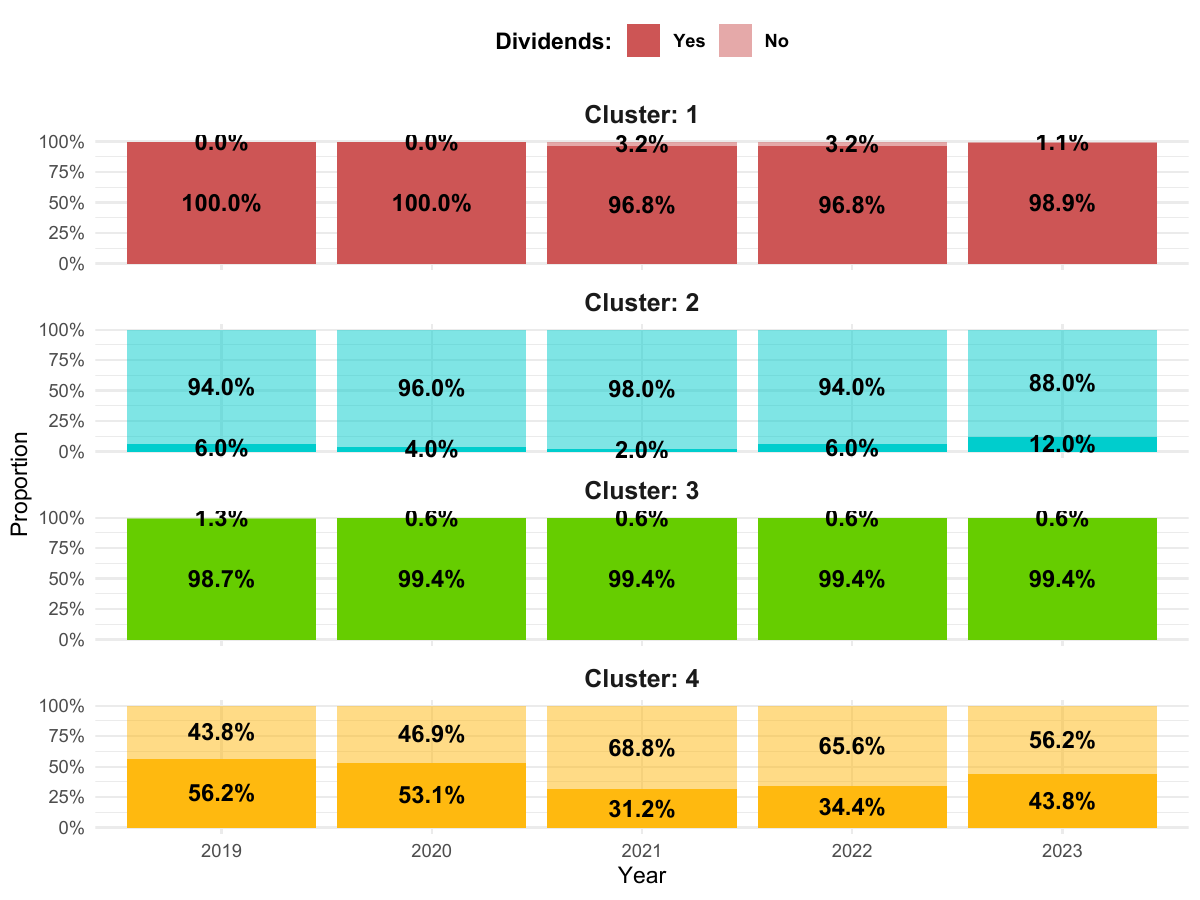}
  \caption{Dividends}
  \label{fig:Dividends}
\end{subfigure}%
\hspace*{.2in}
\begin{subfigure}{.5\textwidth}
  \centering
  \includegraphics[width=1\linewidth]{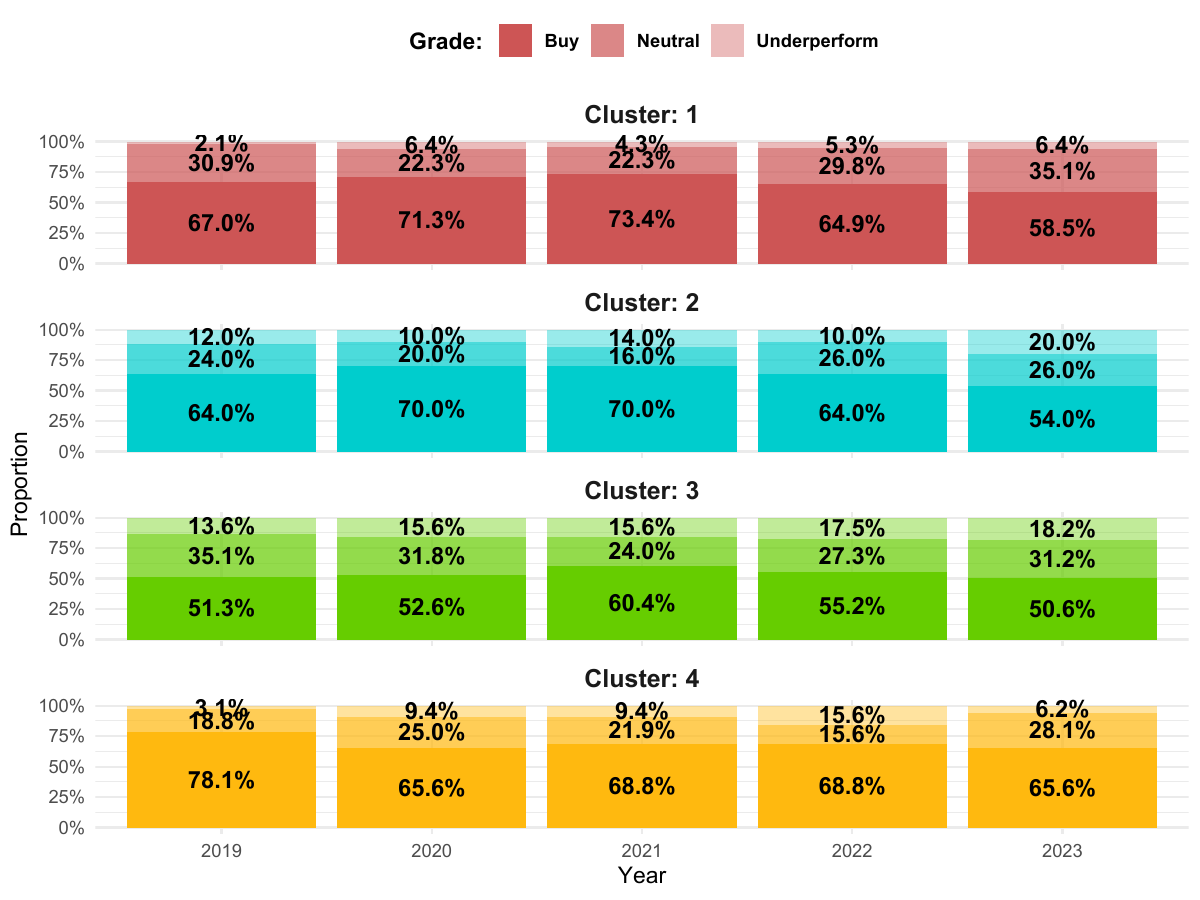}
  \caption{Grades}
  \label{fig:Grades}
\end{subfigure}
\caption{Observed variables values for each cluster. Note that for graphical reason in plots (a) and (b) the company NVIDIA has been removed from the set, due to its out-of-scale values compared to the others companies.}
\label{fig:VarMeans}
\end{figure}

\begin{itemize}
\item \textbf{Cluster 4}: 32 units.
\begin{itemize}
    \item \textbf{Means}: the cluster is qualified by generally constant strong values for LogReturn, with the exepction of 2022, where the cluster has the lowest negative value. The cluster also has the second highest values for Grade and the highest values for Volume. The values for Dividend are small and fluctuate around zero in time suggesting heterogeneity in the cluster regarding this variable.
    \item \textbf{Correlation in time}: the cluster is characterized the second strongest correlations overall. 
    \item \textbf{Correlation among variables}: the main variable of the cluster concerning variables correlation is the absence of a negative correlation between volume and LogReturn, while a weak negative correlation between Dividend and LogReturn is estimated. 
\end{itemize}
Cluster 4 is defined by its high value of the variable Volume compared to the others. The values of LogReturn are more stable in time, except for 2022. The value of Dividens float around zero, and Figure \ref{fig:VarMeans} shows us that the dividend distribution is almost evenly split for most of the years.\\
It is also the only cluster to have a negative correlation between Dividend and LogReturn, implying that stocks with higher returns are also the ones with no dividends. This paradox can be explained by looking at the sector distribution in Figure \ref{fig:VarMeans} : a majority of the companies whose stocks are allocated to Cluster 4 belong to sectors such as ``Technology" and ``Consumer Cyclical", and when we look at Table \ref{table:stocks} we realize it includes companies like Amazon, Tesla, Netflix, Nvidia, AMD and Moderna, that do not allocate dividends but prefer to reinvest their profit in R\&D.\\
\end{itemize}

\section{Conclusions}
In this work we have presented a novel approach for modeling longitudinal mixed-type data with unobserved heterogeneity. The model presented does not require the conditional independence assumption. The matrix-variate structure allows for a more parsimonious modeling of multivariate longitudinal data than other models in the literature. Also, it can explicitly model the temporal structure and the association among the responses, that can vary among clusters. An MCMC-EM algorithm to perform inference has been proposed and described. The efficacy of the algorithm has been tested on synthetic data under different sample sizes and different noise ratios. We proved the goodness of this framework to cluster longitudinal mixed-type data and to get clusters that are easy to interpret and to work with even by non-statisticians in a real-world example.\\

However, the proposed model has limitations, focusing only on basic matrix-normal structures.
While considerably more parsimonious than a mixture of multivariate normal distributions, the model seems sensitive to small sample sizes, since, as the number of clusters increases, the number of parameters to estimate can still became troublesome. To improve this aspect, the covariance matrices can be further decomposed to obtain more flexible and parsimonious models, as done for example in \cite{Anderlucci2015Jun} and in \cite{Sarkar2020Feb}. Another solution to this problem can be the one proposed by \cite{Cappozzo2024Dec}.\\
Similarly, the matrix-variate structure is not just inherent to multivariate longitudinal data, but can actually be found in many other applications. The MMM model can be employed in such cases as well, with minimal adjustments required.\\
Moreover, EM algorithm can be leveraged to extend the model to deal with incomplete data under the missing at random (MAR).\\

Finally, one could as well think of employing, with proper adjustments, different underlying continuous distributions, such as heavy-tailed \citep{Tomarchio2020Dec}, skewed \citep{gallaugher2018finite,Melnykov2018Sep} or t-Student \citep{dougru2016finite} distributions to endow the clustering model with different desired properties.

\section*{Acknowledgment}
This work has been realised thanks to the financial support provided by Project IADoc@UdL of the University of Lyon and Université Lumière - Lyon 2 as part of the call for ``doctoral contracts in artificial intelligence 2020" (ANR-20-THIA-0007-01).


\printbibliography
\break

\newpage
\begin{appendices}
\label{secAppendix}

\section{E-step computations}
Here we will expand the computations presented in Section \ref{sec:em}.\\

For Equation \ref{eq:Zi}, the matrix-variate expectation related to count data can be computed by defining $z^{\gamma}_i \in \mathbb{R}^{GT \times 1}$ as the vectorized version of $Z^{\gamma}_i$ and computing
\begin{align}
 \hat{m}^{\gamma,(s+1)}_{ik} :&= \mathbb{E}(z^{\gamma}_i |\ell_{ik} = 1,\mathbf{Y} ,\hat{\boldsymbol{\Theta}}^{(s)}) = \\ \nonumber
 &= \int_{\mathbb{R}} z^{\gamma}_i \cdot \frac{\prod_{t}^{T}\prod_{g}^{G} \mathcal{P}(y^{\gamma}_{igt}|\exp(z^{\gamma}_{igt}))\cdot \mathcal{MN}_{GT}(z^{\gamma}_i|\text{vec}(M^{(s),\gamma|\alpha,\beta}_k),\Sigma^{(s),\gamma|\alpha,\beta}_k \otimes \Phi^{(s)}_k)}{\int_{\mathbb{R}} \prod_{t}^{T}\prod_{g}^{G} \mathcal{P}(y^{\gamma}_{igt}|\exp(z^{\gamma}_{igt}))\cdot \mathcal{MN}_{GT}(z^{\gamma}_i|\text{vec}(M^{(s),\gamma|\alpha,\beta}_k),\Sigma^{(s),\gamma|\alpha,\beta}_k \otimes \Phi^{(s)}_k) d z^{\gamma}_i} \, d z^{\gamma}_i.
\end{align}
This integral does not have any close form solution, so we resort to numerically compute it through the No-U-Turn sampler implemented in the R package \texttt{Rstan}.\\

Then, $\hat{M}^{\gamma,(s+1)}_{ik} := \text{vec}_{G \times T}^{-1}(\hat{m}^{\gamma,(s+1)}_{ik})$, $\text{vec}_{G \times T}^{-1}$ being the inverse of the vectorization function, i.e. the function mapping from a $GT$-dimensional vector to a $O \times T$ matrix.\\

The matrix-variate expectation related to categorical data can be computed by defining $z^{\beta}_i \in \mathbb{R}^{OT \times 1}$ as the vectorized version of $Z^{\beta}_i$ and computing
\begin{equation}
 \hat{m}^{\beta,(s+1)}_{ik} := \mathbb{E}(z^{\beta}_i |\ell_{ik} = 1,\mathbf{Y} ,\hat{\boldsymbol{\Theta}}^{(s)}) = \int_{\Omega_r} z^{\beta}_i \mathcal{MN}_{OT}(z^{\beta}_i|\text{vec}(M^{(s),\beta|\alpha}_k),\Sigma^{(s),\beta|\alpha}_k \otimes \Phi^{(s)}_k) d z^{\beta}_i
\end{equation}
through the use of a Gibbs sampler to sample from a truncated multivariate normal distribution.\\

Then, as we did for count data; we map the estimated values back to a matrix form as $\hat{M}^{\beta,(s+1)}_{ik} := \text{vec}_{O \times T}^{-1}(\hat{m}^{\beta,(s+1)}_{ik})$.\\

For Equation \ref{eq:ZiPhi}, to compute $D^{(s)}_{ik}$, we start by defining $\hat{\varphi}^{(s)}_{k,gd}$ as the $(g,d)^{th}$ element of  $\hat{\Phi}^{-1(s)}_k$. Then, the $(h,t)^{th}$ element of $Z^{\beta}_i\Phi^{-1}_k Z_i^{\beta \top}$ would be $\sum_{d=1}^T \sum_{g=1}^T z^{\beta}_{i,hg} \hat{\varphi}^{(s)}_{k,gd} z^{\beta}_{i,td}$ and we would get

\begin{align}
  \hat{D}^{(s)}_{ik} &:= \mathbb{E}(Z^{\beta}_i\Phi^{-1}_k Z_i^{\beta \top}|\ell_{ik} = 1,\hat{\boldsymbol{\Theta}}^{(s)},\mathbf{Y})) = \nonumber \\
   &= \left(\sum_{d=1}^T \sum_{g=1}^T \hat{S}^{\beta,(s+1)}_{ik,[(g-1)O+h,(d-1)O+t]}\hat{\varphi}^{(s)}_{k,gd} \right)_{h,t},
   \label{eq:d}
\end{align}

where we make use of the the elements of 
\begin{equation}
 \hat{S}^{\beta,(s+1)}_{ik} := \mathbb{E}(z^{\beta}_i z_i^{\beta \top}|\ell_{ik} = 1,\mathbf{Y} ,\hat{\boldsymbol{\Theta}}^{(s)}) = \int_{\Omega_r} z^{\beta}_i z_i^{\beta \top} \mathcal{MN}_{OT}(z^{\beta}_i|\text{vec}(M^{(s),\beta|\alpha}_k),\Sigma^{(s),\beta|\alpha}_k \otimes \Phi^{(s)}_k) d z^{\beta}_i.
\end{equation}

The samples generated to calculate the first moment $m^{\beta, (s+1)}_{ik}$ can be reused to compute the matrix $\hat{S}_{ik}^{(s+1)}$, that can be approximated by calculating the inner product of the vectors used to compute $m^{\beta, (s+1)}_{ik}$ then calculating the sample mean of these inner products.\\

Similarly, for $\hat{B}^{(s)}_{ik}$ the $(h,t)^{th}$ element of $Z^{\gamma}_i\Phi^{-1}_k Z_i^{\gamma \top}$ would be $\sum_{d=1}^T \sum_{g=1}^T z^{\gamma}_{i,hg} \varphi_{k,gd} z^{\gamma}_{i,td}$ and we would get

\begin{align}
  \hat{B}^{(s)}_{ik} &:= \mathbb{E}(Z^{\gamma}_i\Phi^{-1}_k Z_i^{\gamma \top}|\ell_{ik} = 1,\hat{\boldsymbol{\Theta}}^{(s)},\mathbf{Y})) = \nonumber \\
   & = \left(\sum_{d=1}^T \sum_{g=1}^T \hat{S}^{\gamma,(s+1)}_{ik,[(g-1)G+h,(d-1)G+t]}\hat{\varphi}^{(s)}_{k,gd} \right)_{h,t},
   \label{eq:b}
\end{align}

where we make use of the the elements of 
\begin{align}
 \hat{S}^{\gamma,(s+1)}_{ik} :&= \mathbb{E}(z^{\gamma}_i z_i^{\gamma \top}|\ell_{ik} = 1,\mathbf{Y} ,\hat{\boldsymbol{\Theta}}^{(s)}) = \\ \nonumber
 &= \int_{\mathbb{R}} z^{\gamma}_i z_i^{\gamma \top}\cdot \frac{\prod_{t}^{T}\prod_{g}^{G} \mathcal{P}(y^{\gamma}_{igt}|\exp(z^{\gamma}_{igt}))\cdot \mathcal{MN}_{GT}(z^{\gamma}_i|\text{vec}(M^{(s),\gamma|\alpha,\beta}_k),\Sigma^{(s),\gamma|\alpha,\beta}_k \otimes \Phi^{(s)}_k)}{\int_{\mathbb{R}} \prod_{t}^{T}\prod_{g}^{G} \mathcal{P}(y^{\gamma}_{igt}|\exp(z^{\gamma}_{igt}))\cdot \mathcal{MN}_{GT}(z^{\gamma}_i|\text{vec}(M^{(s),\gamma|\alpha,\beta}_k),\Sigma^{(s),\gamma|\alpha,\beta}_k \otimes \Phi^{(s)}_k) d z^{\gamma}_i} \, d z^{\gamma}_i.
\end{align}

As before, the samples generated to calculate the first moment $\hat{m}_{ik}^{\gamma,(s+1)}$ can be reused to compute the matrix $\hat{S}_{ik}^{\gamma,(s+1)}$ by calculating the mean of the inner product between them.\\

Finally, for Equation \ref{eq:ZiSigma}, let us define by $\hat{\sigma}^{(s),\beta\beta}_{k,gd}$ the $(g,d)^{th}$ element of the block $\hat{\Sigma}^{-1(s),\beta \beta}_k$ . Then, the $(h,t)^{th}$ element of $Z_i^{\beta \top}\Sigma^{\beta \beta}_k Z^{\beta}_i$ is $\sum_{d=1}^O \sum_{g=1}^O z_{i,gh} \hat{\sigma}^{(s),\beta\beta}_{k,gd} z_{i,dt}$, and we get

\begin{align}
\hat{C}_{ik}^{(s)} &:=  \mathbb{E}(Z_i^{\beta \top}\Sigma^{\beta \beta}_k Z^{\beta}_i|\ell_{ik} = 1,\hat{\boldsymbol{\Theta}}^{(s)},\mathbf{Y}) = \nonumber\\
&= \left(\sum_{d=1}^O \sum_{g=1}^O \hat{S}^{\beta, (s+1)}_{ik,[(h-1)J+g,(t-1)J+d]}\hat{\sigma}^{(s),\beta\beta}_{k,gd} \right)_{h,t}.
   \label{eq:c}
\end{align}

For $\hat{A}_{ik}^{(s)}$, let indicate by $\hat{\sigma}^{(s),\gamma \gamma}_{k,gd}$ the $(g,d)^{th}$ element of block $\hat{\Sigma}^{-1(s),\gamma \gamma}_k$. Then, the $(h,t)^{th}$ element of $Z_i^{\gamma \top}\Sigma^{\gamma \gamma}_k Z^{\gamma}_i$ is $\sum_{d=1}^O \sum_{g=1}^O z_{i,gh} \hat{\sigma}^{(s),\gamma \gamma}_{k,gd} z_{i,dt}$, and we get

\begin{align}
\hat{A}_{ik}^{(s)} &:=  \mathbb{E}(Z_i^{\gamma \top}\Sigma^{\gamma \gamma}_k Z^{\gamma}_i|\ell_{ik} = 1,\hat{\boldsymbol{\Theta}}^{(s)},\mathbf{Y}) = \nonumber \\
&= \left(\sum_{d=1}^O \sum_{g=1}^O \hat{S}^{\gamma, (s+1)}_{ik,[(h-1)J+g,(t-1)J+d]}\hat{\sigma}^{(s),\gamma \gamma}_{k,gd} \right)_{h,t}.
   \label{eq:a}
\end{align}

\section{Cluster interpretation}
\label{appendix:clust_inter}
In this section interpretations for other clusters referred in Section \ref{sec:interpretation} are given.

\begin{itemize}
\item \textbf{Cluster 1}: 94 units.
    \begin{itemize}
        \item \textbf{Means}: the cluster means show the highest values for Dividend and Grade, and the second highest for Volume. The results for LogReturn are more shaded: the cluster has the lowest mean for 2019, 2020 (the only one negative for that year) and 2023. At the same time, it is the only cluster to have non-negative LogReturns for 2022. 
        \item \textbf{Correlation in time}: the cluster is characterized by a fading and weaker correlations among times than other clusters, especially regarding 2022 to the previous years. 
        \item \textbf{Correlation among variables}: the cluster is characterized by feeble correlations among Returns, Grade and Dividend, yet these correlations are stronger than in other clusters. Some soft negative correlations are estimated between Volume, Grade and Return.
    \end{itemize}
We can describe Cluster 1 as the cluster of more ``traditional" stocks. Stocks belonging to this cluster have good grades, usually grant dividends and are among the most exchanged, ensuring good liquidity.\\
By looking at Figure \ref{fig:comp_C1}, we can notice that the cluster is the ones with more variety of composing sectors. This might explain why it is the only cluster that  experienced a fall in LogReturns in 2020, at the height of the COVID-19 pandemic and of the consequent lock-downs, which had a major impact on more traditional sectors. Figure \ref{fig:VarMeans} suggests that indeed the stocks gave right to dividends even for the entirety of them in 2019 and 2020. We can also point out to the fact that during 2020 and 2021 the percentage of stocks marked as ``Buy" for this cluster increased, probably in view of the end of toughest pandemic period and in light of the lower prices of the stocks. The grades distribution changes during 2022 and 2023 mostly in favor of ``Neutral". The correlations among times suggest that the behaviour is less constant in time with respect of the other clusters. Moreover, the negative correlations between Volume, LogReturns and Grade may indicate that the increase in volume exchange is generally related to selling, as the volume increase when grades and returns decreases. 

\item \textbf{Cluster 2}: 50 units.
\begin{itemize}
    \item \textbf{Means}: the cluster has the highest means regarding LogReturns for the first two years and the second most important negative value for 2022. It is the only cluster with relatively strong negative values for Dividend. It has also the lowest values for Volume. 
    \item \textbf{Correlation in time}: it is the cluster with the strongest positive correlation in time.
    \item \textbf{Correlation variables}: the cluster is characterized by the presence of weak correlation between LogReturn, Grade ans Dividend, and of a weak negative correlation between Volume and LogReturn.
\end{itemize}
Cluster 2 has the main characteristics to be the only cluster with negative values for Dividend. A look at Figure \ref{fig:VarMeans} shows us that indeed that almost none of the stocks allocated to the cluster gave right to a dividend, a situation that slightly improves in 2023. The low values for Volume compared to the other clusters indicate that the stocks in this cluster are among the less exchanged. The grades distribution show that there is a high percentage of stocks marked as ``Buy" until 2021, but it decreases and in 2023 the cluster has the highest percentage of stocks marked as ``Underperform". 2022 appears to be a bad year for the stocks belonging to the cluster, but with the expect of this year the cluster has the most stable values for LogReturn. The sector composition of the cluster shows a dominance of the sectors ``Healthcare" and ``Technology", which might explain the good performance during the pandemic, as these sectors were among the ones to actually profit during the pandemic. The same reason might explain the 2022 performance, where staff lay-offs and decrease in investments due to over-investments during the pandemics hit particularly the IT sector.

\item \textbf{Cluster 3}: 154 units.
\begin{itemize}
    \item \textbf{Means}: the cluster has the second highest means for LogReturns for 2019 and 2021, and the lowest negative value for 2022. It has the second highest values for Dividend and the second smallest values for Volume. 
    \item \textbf{Correlation in time}: the cluster has the overall strong positive correlations in time. 
    \item \textbf{Correlation among variables}: the cluster is mainly characterized by the weak negative correlation between Volume and LogReturn, and the absence of other meaningful correlations.
\end{itemize}
Cluster 3 can be seen as cluster between Cluster 1 and Cluster 2: both Volume and Grade have values in between the two, and the same can be almost be said for LogReturn. The main exception to this description is Dividend, since for Cluster 3 the values are high, and if we look at Figure \ref{fig:VarMeans} almost 100\% of the stocks gave right to a dividend. Besides, the percentage of stocks releasing dividends is surprisingly stable over time. \\
Moreover, concerning the variables Grade, the cluster is the one with the smallest percentage of stocks classified as ``Buy", while it has the highest percentage of stocks marked as ``Neutral" among all the clusters.\\
Its main sector is ``Industrials", but we can see from Figure \ref{fig:cluster_sectors} that its composition is diversified, more like Cluster 1 than Cluster 2.
\end{itemize}

\section{Simulations}
\label{secSimulations}
\setcounter{table}{0}
\renewcommand{\thetable}{C\arabic{table}}
\setcounter{figure}{0}
\renewcommand{\thefigure}{C\arabic{figure}}

\begin{longtable}{crrr}
\caption{Means matrices for simulation}
\label{table::simul_means}\\
\hline 
\endfirsthead
\textbf{Cluster 1} & T1 & T2 & T3 \\
 \hline
 V1 & 1.75 & 1.75 & 1.75 \\
 V2 & 1.75 & 1.75 & 1.75\\
 V3 & -0.25 & -0.25 & -0.25\\
 V4 & 1 & 1 & 1 \\
 \space
 \hline
 \hline
 \textbf{Cluster 2} & T1 & T2 & T3 \\
 \hline
 V1 & 2.75 & 2.75 & 2.75 \\
 V2 & 2.75 & 2.75 & 2.75\\
 V3 & 0.25 & 0.25 & 0.25\\
 V4 & 2.5 & 2.5 & 2.5 \\
\hline
\end{longtable}

\section{Real data}
\label{appendix:real}
\setcounter{table}{0}
\renewcommand{\thetable}{D\arabic{table}}
\setcounter{figure}{0}
\renewcommand{\thefigure}{D\arabic{figure}}

\begin{longtable}{crrrrr}
\caption{Clusters' means over time. The estimated parameter $\hat{\pi}$ = (0.287, 0.156, 0.460, 0.096)} \\
\label{table:means}\\
\hline
\textbf{Cluster} & \textbf{2019} & \textbf{2020} & \textbf{2021} & \textbf{2022} & \textbf{2023} \\
\hline
\endfirsthead

\multicolumn{6}{c}%
{{\bfseries Table \thetable{} -- continued from previous page}} \\
\hline
\textbf{Cluster} & \textbf{2019} & \textbf{2020} & \textbf{2021} & \textbf{2022} & \textbf{2023} \\
\hline
\endhead

\hline
\endlastfoot

\multicolumn{6}{l}{\textbf{Cluster 1}} \\
\hline
Return  &  19.77 & -3.34 & 28.72 & 0.03 & 5.07  \\
Grade  &  3.93 & 4.16 & 4.07 & 4.07 & 3.71  \\
Dividend  &  4.07 & 4.04 & 3.44 & 3.51 & 3.58  \\
Volume  &  7.35 & 7.59 & 7.38 & 7.44 & 7.33  \\
\hline
\multicolumn{6}{l}{\textbf{Cluster 2}} \\
\hline
Return  &  34.69 & 31.77 & 26.73 & -27.7 & 22.21  \\
Grade  &  3.00 & 3.24 & 3.20 & 3.04 & 2.69  \\
Dividend  &  -1.57 & -1.92 & -2.05 & -2.00 & -1.68  \\
Volume  &  5.72 & 5.82 & 5.52 & 5.70 & 5.66  \\
\hline
\multicolumn{6}{l}{\textbf{Cluster 3}} \\
\hline
Return  &  27.92 & 9.54 & 28.06 & -10.87 & 12.34  \\
Grade  &  3.09 & 3.19 & 3.44 & 3.23 & 3.08  \\
Dividend  &  3.34 & 3.65 & 3.58 & 3.81 & 4.01  \\
Volume  &  6.02 & 6.15 & 5.91 & 6.00 & 5.98  \\
\hline
\multicolumn{6}{l}{\textbf{Cluster 4}} \\
\hline
Return  &  21.29 & 22.72 & 24.87 & -43.95 & 36.45  \\
Grade  &  4.52 & 3.71 & 3.71 & 3.52 & 3.46  \\
Dividend  &  0.50 & 0.38 & -0.88 & -0.54 & -0.13  \\
Volume  &  8.04 & 8.84 & 8.44 & 8.52 & 8.33  \\
\end{longtable}

\begin{longtable}{crrrrr}
\caption{Clusters' time covariances}
\label{table:Phi}\\
  \hline
 \textbf{Cluster 1}& 2019 & 2020 & 2021 & 2022 & 2023 \\ 
  \hline
\hline
2019  &  1.36 & 0.94 & 0.75 & 0.61 & 0.66  \\
2020  &  0.94 & 1.64 & 0.81 & 0.61 & 0.67  \\
2021  &  0.75 & 0.81 & 1.51 & 0.99 & 0.77  \\
2022  &  0.61 & 0.61 & 0.99 & 1.68 & 0.83  \\
2023  &  0.66 & 0.67 & 0.77 & 0.83 & 1.3  \\
\space
  \hline
 \textbf{Cluster 2}& 2019 & 2020 & 2021 & 2022 & 2023 \\ 
 \hline
2019  &  2.25 & 1.97 & 1.81 & 1.69 & 1.79  \\
2020  &  1.97 & 2.42 & 1.94 & 1.78 & 1.92  \\
2021  &  1.81 & 1.94 & 2.54 & 2.02 & 2.07  \\
2022  &  1.69 & 1.78 & 2.02 & 2.69 & 2.2  \\
2023  &  1.79 & 1.92 & 2.07 & 2.2 & 2.73  \\  
  \hline
 \textbf{Cluster 3}& 2019 & 2020 & 2021 & 2022 & 2023 \\ 
  \hline
2019  &  1.63 & 1.28 & 1.2 & 1.06 & 1.06  \\
2020  &  1.28 & 1.81 & 1.3 & 1.09 & 1.09  \\
2021  &  1.2 & 1.3 & 1.93 & 1.29 & 1.27  \\
2022  &  1.06 & 1.09 & 1.29 & 1.95 & 1.29  \\
2023  &  1.06 & 1.09 & 1.27 & 1.29 & 1.9  \\
  \hline
  \textbf{Cluster 4}& 2019 & 2020 & 2021 & 2022 & 2023 \\ 
  \hline
22019  &  2.61 & 1.57 & 1.59 & 1.5 & 1.61  \\
2020  &  1.57 & 2.06 & 1.42 & 1.16 & 1.27  \\
2021  &  1.59 & 1.42 & 1.95 & 1.48 & 1.48  \\
2022  &  1.5 & 1.16 & 1.48 & 1.92 & 1.52  \\
2023  &  1.61 & 1.27 & 1.48 & 1.52 & 1.97  \\
  \hline
\end{longtable}

\begin{longtable}{crrrr}
\caption{Clusters' variables covariances}
\label{table:Sigma}\\
  \hline
 \textbf{Cluster 1} & Return & Grade & Dividend & Volume \\
  \hline
\hline
Return  &  589.69 & 6.4 & 6.22 & -0.87  \\
Grade  &  6.4 & 3.36 & 0.92 & -0.06  \\
Dividend  &  6.22 & 0.92 & 3.74 & -0.03  \\
Volume  &  -0.87 & -0.06 & -0.03 & 0.1  \\
\space
  \hline
 \textbf{Cluster 2} & Return & Grade & Dividend & Volume \\ 
 \hline
Return  &  976.02 & 5.09 & 3.85 & -1.82  \\
Grade  &  5.09 & 2.02 & 0.54 & 0  \\
Dividend  &  3.85 & 0.54 & 1.98 & -0.01  \\
Volume  &  -1.82 & 0 & -0.01 & 0.15  \\  
  \hline
 \textbf{Cluster 3} & Return & Grade & Dividend & Volume \\ 
  \hline
Return  &  521.79 & 1.15 & 1.89 & -0.91  \\
Grade  &  1.15 & 3.6 & 0.97 & -0.01  \\
Dividend  &  1.89 & 0.97 & 3.89 & -0.01  \\
Volume  &  -0.91 & -0.01 & -0.01 & 0.07  \\
  \hline
  \textbf{Cluster 4} & Return & Grade & Dividend & Volume \\ 
  \hline
Return  &  2378.8 & 4.74 & -8.1 & 1.2  \\
Grade  &  4.74 & 2.64 & 0.61 & -0.05  \\
Dividend  &  -8.1 & 0.61 & 2.31 & 0.04  \\
Volume  &  1.2 & -0.05 & 0.04 & 0.32  \\
  \hline
\end{longtable}

\begin{figure}[!ht]
\hspace*{-1.5in}
\begin{subfigure}{.5\textwidth}
  \centering
  \includegraphics[width=1\linewidth]{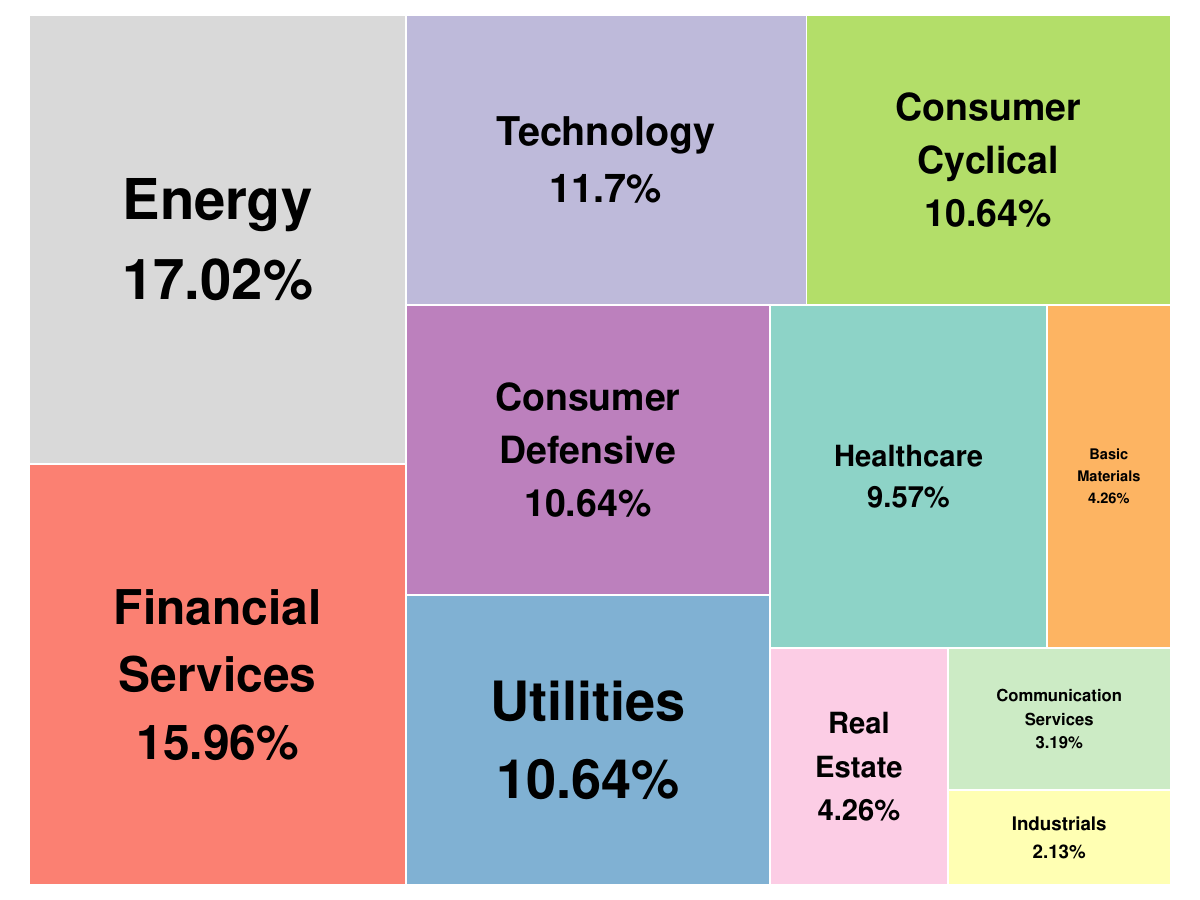}
  \caption{Cluster 1}
  \label{fig:comp_C1}
\end{subfigure}%
\hspace*{-.8in}
\begin{subfigure}{.5\textwidth}
  \centering
  \includegraphics[width=1\linewidth]{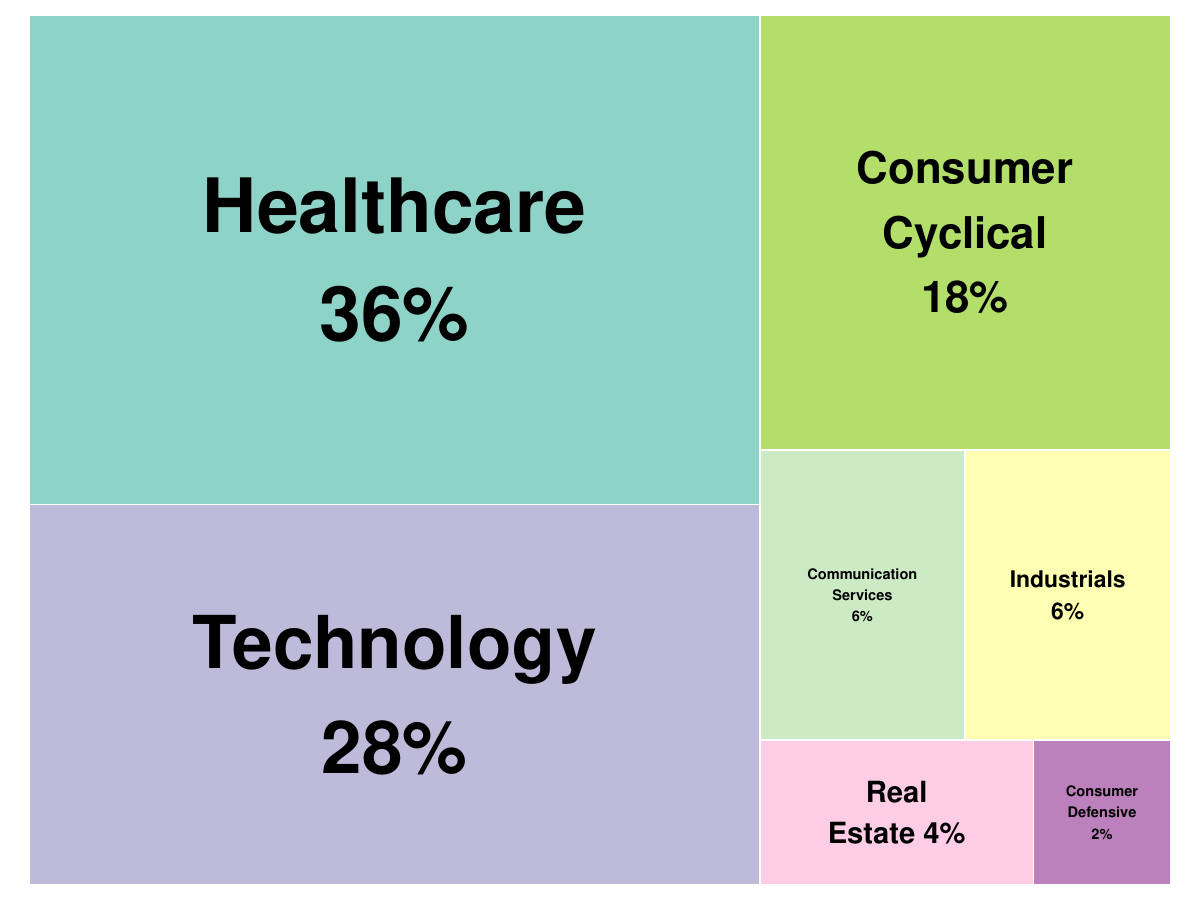}
  \caption{Cluster 2}
  \label{fig:comp_C2}
\end{subfigure}
\hspace*{-.4in}
\begin{subfigure}{.5\textwidth}
  \centering
  \includegraphics[width=1\linewidth]{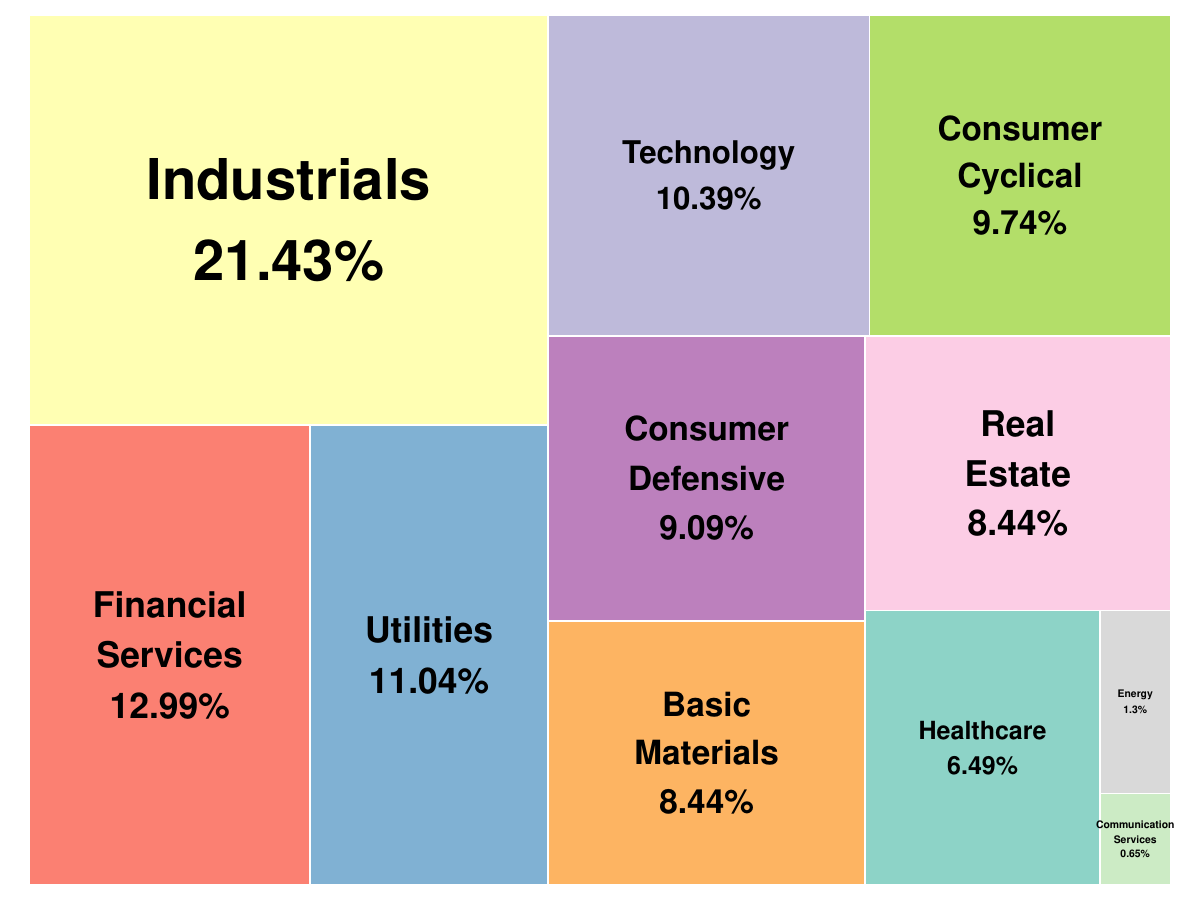}
  \caption{Cluster 3}
  \label{fig:comp_C3}
\end{subfigure}%
\hspace*{.3in}
\begin{subfigure}{.5\textwidth}
  \centering
  \includegraphics[width=1\linewidth]{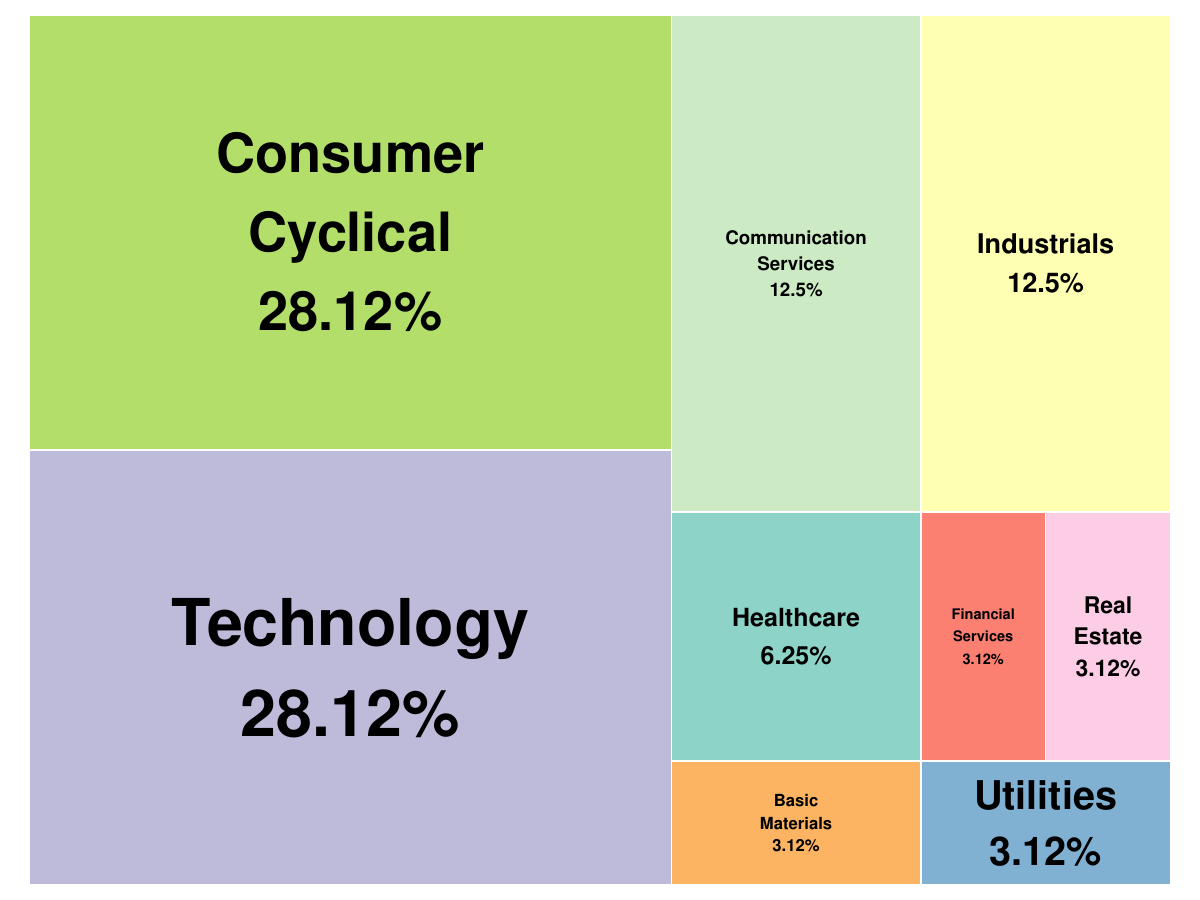}
  \caption{Cluster 2}
  \label{fig:comp_C4}
\end{subfigure}
\caption{Clusters’ sectors composition}
\label{fig:cluster_sectors}
\end{figure}

\begin{figure}[ht!]
\centering
\includegraphics[scale = .8]{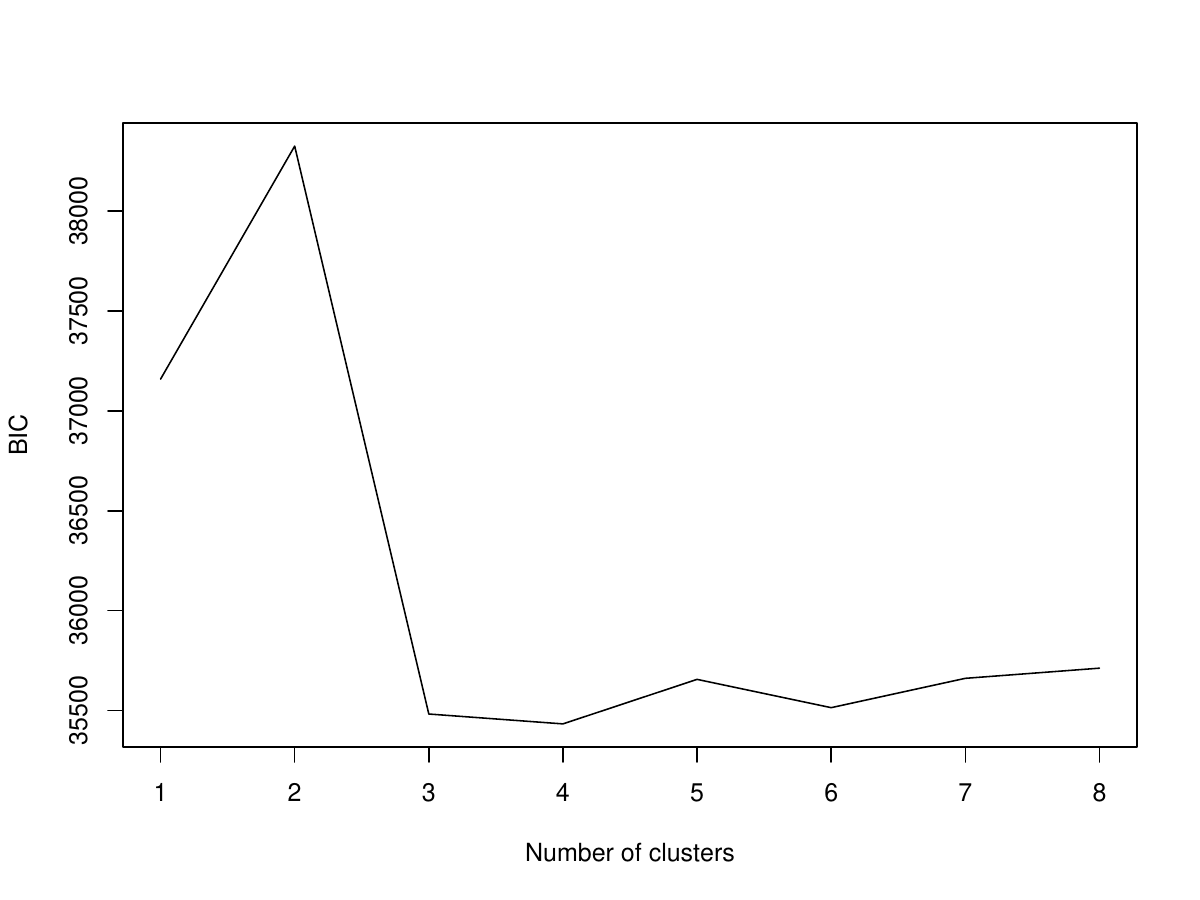}
\caption{Visualization of BIC for K as results of application on real data. Kmeans++ initialization.}
\label{fig:BIC_realdata_viz}
\end{figure}

\begin{longtable}[c]{llll}
\caption{Stocks' tickers in each cluster}
\label{table:stocks}\\
\hline
Cluster 1 & Cluster 2 & Cluster 3 & Cluster 4 \\
\hline
\endfirsthead
\hline
Cluster 1 & Cluster 2 & Cluster 3 & Cluster 4 \\
\hline
\endhead
ABBV, AES, AIG & ADBE, ADSK, ANET & A, ACGL, ACN & AAL, AAPL, AMD  \\
AMAT, BK, BKR & APTV, AZO, BKNG & ADI, ADP, AEE & AMZN, AVGO, BA  \\
BMY, BX, CFG & BSX, CBRE, CDNS & ALB, ALL, AME & CMG, CRWD, CZR  \\
CL, CMCSA, CNP & CNC, CRL, CSGP & APD, AVB, AVY & DAL, DIS, EXPE  \\
COP, CSCO, CSX & CTLT, DECK, DLTR & AWK, AXP, BALL & F, FCX, GM  \\
CVS, CVX, D & DVA, DXCM, EPAM & BAX, BBY, BEN & GOOGL, INTC, MRNA  \\
DD, DOW, DVN & EW, FFIV, FSLR & BWA, BXP, CAT & MSFT, NCLH, NFLX  \\
EBAY, EOG, EXC & FTNT, GNRC, HOLX & CBOE, CDW, CE & NVDA, PCG, PFE  \\
FANG, FE, FIS & HSIC, IDXX, IQV & CF, CHD, CHRW & PYPL, RCL, SPG  \\
FITB, FOXA, GILD & ISRG, IT, KMX & CLX, CME, CMI & T, TSLA, UAL  \\
GLW, HAL, HBAN & LH, LULU, MHK & CMS, COST, CPB & UBER, WDC  \\
HD, HPE, HPQ & MOH, MTCH, MTD & CPT, CTAS, CTSH &   \\
IBM, IVZ, JPM & NOW, NVR, ORLY & DE, DFS, DGX &   \\
KDP, KHC, KIM & PANW, PAYC, PTC & DHI, DOV, DPZ &   \\
KMI, KR, LLY & QRVO, TDG, TMUS & DRI, DTE, DUK &   \\
LOW, LUV, LVS & TTWO, URI, VRTX & EA, ED, EFX &   \\
MCHP, MDLZ, MDT & WAT, WST & EIX, EL, ELV &   \\
MET, MGM, MO &  & EMN, EMR, EQR &   \\
MOS, MPC, MRK &  & ES, ESS, ETN &   \\
MRO, NEE, NEM &  & ETR, EVRG, EXR &   \\
NI, NKE, O &  & FDS, FDX, FMC &   \\
OKE, ORCL, OXY &  & FTV, GD, GPC &   \\
PEP, PG, PM &  & GS, HCA, HES &   \\
PPL, QCOM, RF &  & HII, HON, HRL &   \\
SBUX, SCHW, SLB &  & HSY, HUM, ICE &   \\
SO, SYF, TGT &  & INTU, IP, IRM &   \\
TJX, TPR, TXN &  & ITW, JBHT, JBL &   \\
UNH, USB, V &  & JNPR, K, KKR &   \\
VICI, VLO, VST &  & KLAC, KMB, LEN &   \\
VZ, WBA, WFC &  & LMT, LNT, LRCX &   \\
WMB, WY, WYNN &  & LW, LYB, MA &   \\
XOM &  & MAS, MCD, MCK &   \\
 &  & MLM, MMC, MMM &   \\
 &  & NDAQ, NRG, NSC &   \\
 &  & NTAP, NTRS, NUE &   \\
 &  & NXPI, ODFL, PAYX &   \\
 &  & PCAR, PEG, PH &   \\
 &  & PHM, PKG, PLD &   \\
 &  & PNC, PNW, PPG &   \\
 &  & PSA, RL, ROK &   \\
 &  & RSG, SBAC, SHW &   \\
 &  & SNA, SRE, STLD &   \\
 &  & STT, STZ, SWK &   \\
 &  & SWKS, SYY, TAP &   \\
 &  & TER, TMO, TRGP &   \\
 &  & TROW, TRV, TSCO &   \\
 &  & TSN, TXT, UDR &   \\
 &  & UHS, UNP, UPS &   \\
 &  & VMC, VRSK, VTR &   \\
 &  & WAB, WEC, WELL &   \\
 &  & WM, WRB, XEL &   \\
 &  & ZTS &   \\
\hline
\end{longtable}

\end{appendices}

\end{document}